%% file: aaai25.tex
\documentclass[letterpaper]{article} 
\usepackage{aaai25}  
\usepackage{times}  
\usepackage{helvet}  
\usepackage{courier}  
\usepackage[hyphens]{url}  
\usepackage{graphicx} 
\usepackage{natbib}  
\usepackage{caption} 
\usepackage{algorithm}
\usepackage{algorithmic}

\input{section/macro}
\usepackage{newfloat}
\usepackage{listings}
\DeclareCaptionStyle{ruled}{labelfont=normalfont,labelsep=colon,strut=off} 
\floatstyle{ruled}
\newfloat{listing}{tb}{lst}{}
\floatname{listing}{Listing}
\title{\textit{More Text, Less Point}: Towards 3D Data-Efficient Point-Language Understanding}

\input{section/00_authors}

\usepackage{bibentry}

\begin{document}

\maketitle

\input{section/01_abs}

\input{section/02_intro}
\input{section/03_related_work}
\input{section/04_method}
\input{section/05_experiment}

\input{section/06_conclusion}
\input{section/07_ack}

\bibliography{aaai25}

\clearpage
\input{section/10_appendix}


\end{document}

%% file: section/macro.tex
\usepackage{xspace}
\newcommand{\ours}{GreenPLM\xspace}

\usepackage{color}

\usepackage{microtype}
\usepackage{graphicx}
\usepackage{subfigure}
\usepackage{booktabs} 

\usepackage{amsmath}

\usepackage{amssymb}
\usepackage{mathtools}
\usepackage{amsthm}

\usepackage[capitalize]{cleveref}

\theoremstyle{plain}

\theoremstyle{definition}

\theoremstyle{remark}

\usepackage{microtype}
\usepackage{graphicx}
\usepackage{subfigure}
\usepackage{booktabs} 

\usepackage{multirow}
\usepackage{amsthm,amsmath,amssymb,lipsum}
\usepackage{mathrsfs}
\usepackage{enumerate}
\usepackage{colortbl}
\usepackage{enumitem}
\usepackage{bbding}
\usepackage{pifont}
\usepackage{wasysym}
\usepackage{textcomp}

\usepackage{color}

\definecolor{pretty-blue}{RGB}{0, 113, 188}
\definecolor{linecolor}{gray}{.91} 
\definecolor{linecolor2}{gray}{.95} 
\definecolor{linecolor1}{gray}{.97} 

\usepackage{listings}

\usepackage{multirow}
\usepackage[table,xcdraw]{xcolor}
\usepackage{soul}

 \usepackage{adjustbox}

\usepackage{graphicx}  
\usepackage{subcaption} 
\usepackage{bm}

 \usepackage{graphicx}
\usepackage{amsmath}
\usepackage{amssymb}
\usepackage{booktabs}
\usepackage{multirow}
\usepackage{booktabs, multirow} 
\usepackage{soul}
\usepackage{changepage,threeparttable}
\usepackage[accsupp]{axessibility}


%% file: section/00_authors.tex
\author {
Yuan Tang\textsuperscript{\rm 1}$^,$\equalcontrib,
Xu Han\textsuperscript{\rm 1}$^,$\equalcontrib,
Xianzhi Li\textsuperscript{\rm 1, \rm 2}$^,$\thanks{Corresponding author.},
Qiao Yu\textsuperscript{\rm 1},
Jinfeng Xu\textsuperscript{\rm 1},\\
Yixue Hao\textsuperscript{\rm 1,\rm 3},
Long Hu\textsuperscript{\rm 1,\rm 3},
Min Chen\textsuperscript{\rm 4,\rm 5 }
}
\affiliations {
    \textsuperscript{\rm 1}Huazhong University of Science and Technology, \textsuperscript{\rm 2}Guangdong Intelligent Robotics Institute\\
    \textsuperscript{\rm 3}Guangdong HUST Industrial Technology Research Institute, \textsuperscript{\rm 4}South China University of Technology, \textsuperscript{\rm 5}Pazhou Laboratory\\
    \{yuan\_tang,  xhanxu, xzli, qiaoyu\_epic, yixuehao, hulong\}@hust.edu.cn,  jinfengxu.edu@gmail.com, minchen@ieee.org
}

%% file: section/01_abs.tex
\begin{abstract}
Enabling Large Language Models (LLMs) to comprehend the 3D physical world remains a significant challenge. Due to the lack of large-scale 3D-text pair datasets, the success of LLMs has yet to be replicated in 3D understanding. In this paper, we rethink this issue and propose a new task: 3D Data-Efficient Point-Language Understanding. The goal is to enable LLMs to achieve robust 3D object understanding with minimal 3D point cloud and text data pairs. To address this task, we introduce GreenPLM, which leverages more text data to compensate for the lack of 3D data. First, inspired by using CLIP to align images and text, we utilize a pre-trained point cloud-text encoder to map the 3D point cloud space to the text space. This mapping leaves us to seamlessly connect the text space with LLMs. Once the point-text-LLM connection is established, we further enhance text-LLM alignment by expanding the intermediate text space, thereby reducing the reliance on 3D point cloud data. Specifically, we generate 6M free-text descriptions of 3D objects, and design a three-stage training strategy to help LLMs better explore the intrinsic connections between different modalities. To achieve efficient modality alignment, we design a zero-parameter cross-attention module for token pooling. Extensive experimental results show that GreenPLM requires only 12\% of the 3D training data used by existing state-of-the-art models to achieve superior 3D understanding. Remarkably, GreenPLM also achieves competitive performance using text-only data.
\begin{links}
\link{Code}{https://github.com/TangYuan96/GreenPLM}
\end{links}

\end{abstract}

%% file: section/02_intro.tex
\section{Introduction}

Recent advancements in large language models (LLMs) have revolutionized natural language processing, demonstrating emergent intelligence and exceptional capabilities in language understanding and generation~\cite{openai2023gpt, qwen2, dubey2024llama, team2023gemini}. 
However, LLMs are \textit{blind} to the 3D physical world because they lack the ability to capture and understand 3D objects. 
Solving this challenging multimodal 3D-language understanding task could benefit many applications, such as autonomous driving, robotics and embodied AI~\cite{driess2023palm, fu2024drive, brohan2023rt}.
\input{images/conv_demo}

\input{images/idea}
Inspired by CLIP~\cite{radford2021learning}, multimodal large language models (MLLMs) can map different modality inputs to a text space closer to LLMs using pre-trained multimodal encoders, enabling LLMs to understand data beyond just language.
Existing 3D point-language models follow a similar approach, applying LLMs to 3D understanding by learning from 3D point-text data pairs~\cite{luo2024scalable,qi2024gpt4point}. 
For example, PointLLM~\cite{xu2023pointllm} and ShapeLLM~\cite{qi2024shapellm} employ pre-trained multimodal point cloud encoders~\cite{xue2024ulip,qi2024shapellm}, mapping the point cloud space to the text space.
This leaves the alignment of point cloud with LLMs to only align the text space with LLMs, which is relatively easier for LLMs.
Finally, they propose to train the 3D-LLMs with large amount of 3D-text data pairs, thus enhancing the LLMs' 3D understanding capabilities.
However, this field remains under-explored.
The primary reason is that training LLMs requires billions of datas, while 3D-text pair data is scarce because 3D data itself is hard to acquire and requires expensive annotations.
Consequently, the scaling law that drives LLMs success are difficult to achieve in the 3D domain, directly limiting the development of 3D foundation models.

In this paper, we revisit the 3D data bottleneck and pose a question: \textbf{\textit{Can we achieve robust 3D understanding with minimal 3D data?}}
To answer this question, we propose a new task: 3D Data-Efficient Point-Language Understanding (3DEPL). 
The goal is to enable LLMs to achieve robust 3D understanding using as little 3D point cloud-text data pairs as possible. 
This requires the model to explore the intrinsic connections between different modalities, and effectively leverage the powerful language comprehension capabilities of LLMs to achieve data-efficient 3D understanding.

To address this data-limited multimodal alignment problem, we propose \ours. 
Intuitively,  as shown in Fig.~\ref{fig:idea}, we observe that after establishing the \textit{point-text-LLM} connection, instead of increasing point-text data pairs to optimize the \textit{point-text} mapping like in existing methods~\cite{xu2023pointllm, qi2024shapellm}, we can also enhance the \textit{text-LLM} alignment by simply adding more text data. 
This approach can also improve the point-LLM alignment and, more importantly, reduce the reliance on point-text data pairs, shifting the data bottleneck from expensive and scarce 3D-text data to abundant and cheap text data.
That is, the text-LLM alignment method fits perfectly with the goal of 3D data-efficient point-language understanding, also offers  an alternative solution for aligning point clouds with LLMs, enabling \ours to achieve robust 3D understanding even with limited 3D data.

In detail, \ours solves the 3DEPL task with key techniques across three perspectives: data, training strategy, and model architecture.
(1) We bring T3D dataset, a 6M text dataset of 3D object descriptions and conversations for free, the largest to our knowledge, to expand the text space for better \textit{text-LLM} alignment and compensate for the scarcity of expensive 3D data. 
(2) We propose a 3-stage training strategy designed to help LLMs better uncover the intrinsic connections between different modalities.
Specifically, we propose a coarse-to-fine training approach, progressing from data to model. The first two stages fine-tune the LLMs with text-only data, while the final stage uses minimal 3D data for further point-LLMs alignment. 
(3) From the architecture's perspective, we design a parameter-free cross-attention module for token pooling, namely 0M-Pooling, which better utilizes the encoder's output tokens, thereby aligning point clouds with LLMs more effectively.
This, we can achieve excellent performance with only an efficient LLM~\cite{abdin2024phi}.
Together, we can complete training in just  26.6 hours using a single 3090 GPU (24GB), leaving opportunities for efficient end-side deployment.

To fairly and reasonably evaluate the models, we introduce a new metric to measure the efficiency of 3D data usage, and establish a new evaluation benchmark based on open-source LLMs.
Experimental results show that our \ours outperforms previous models using only 12\% of the 3D data. It even surpasses GPT4Point (660K)~\cite{qi2024gpt4point} without any 3D data, maintaining extremely 3D data-efficient point-language understanding, which demonstrates the effectiveness of our approach.
The contributions of this paper are as follows:
\begin{itemize}
    \item We introduce a new task of 3D data-efficient point-language understanding, aiming to enable LLMs to achieve robust 3D understanding with minimal 3D data.
    \item We propose \ours to tackle this 3D data-limited task from a novel perspective, enhancing point-LLM alignment with more free-text data. Specifically, we introduce a 6M T3D dataset, design a 3-stage training strategy, and present a 0M-Pooling module for token pooling.
    \item We introduce the Accuracy-to-3D-Data Ratio (A3DR) to measure the efficiency of 3D data usage and establish an evaluation benchmark based on open-source LLMs. 
    \item \ours outperforms previous models using only 12\% of 3D data and even surpasses GPT4Point (660K 3D data) using only text, demonstrating superior 3D data efficiency.
\end{itemize}

%% file: images/conv_demo.tex
 \begin{figure}[h!]
\centering
  \includegraphics[width=\linewidth]{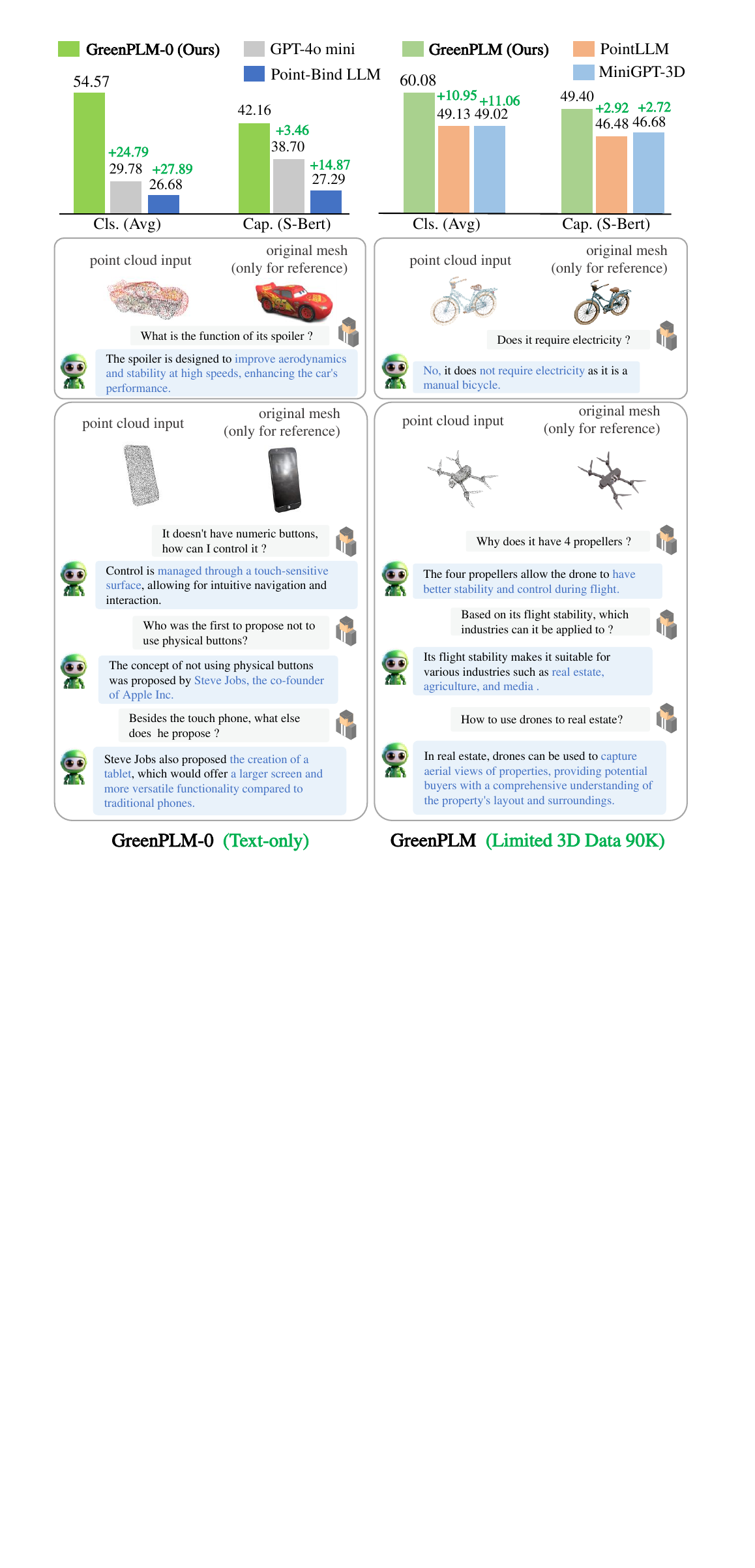}
   
  \caption{
  We propose GreenPLM, which expands the text space to reduce the need for 3D data. 
  GreenPLM achieves strong 3D understanding using just 12\% of the 3D data or even with text-only data.
}
  \label{fig:conv_demo}

\end{figure}

%% file: images/idea.tex
\begin{figure}[t]
    \centering
  \includegraphics[width=\linewidth]{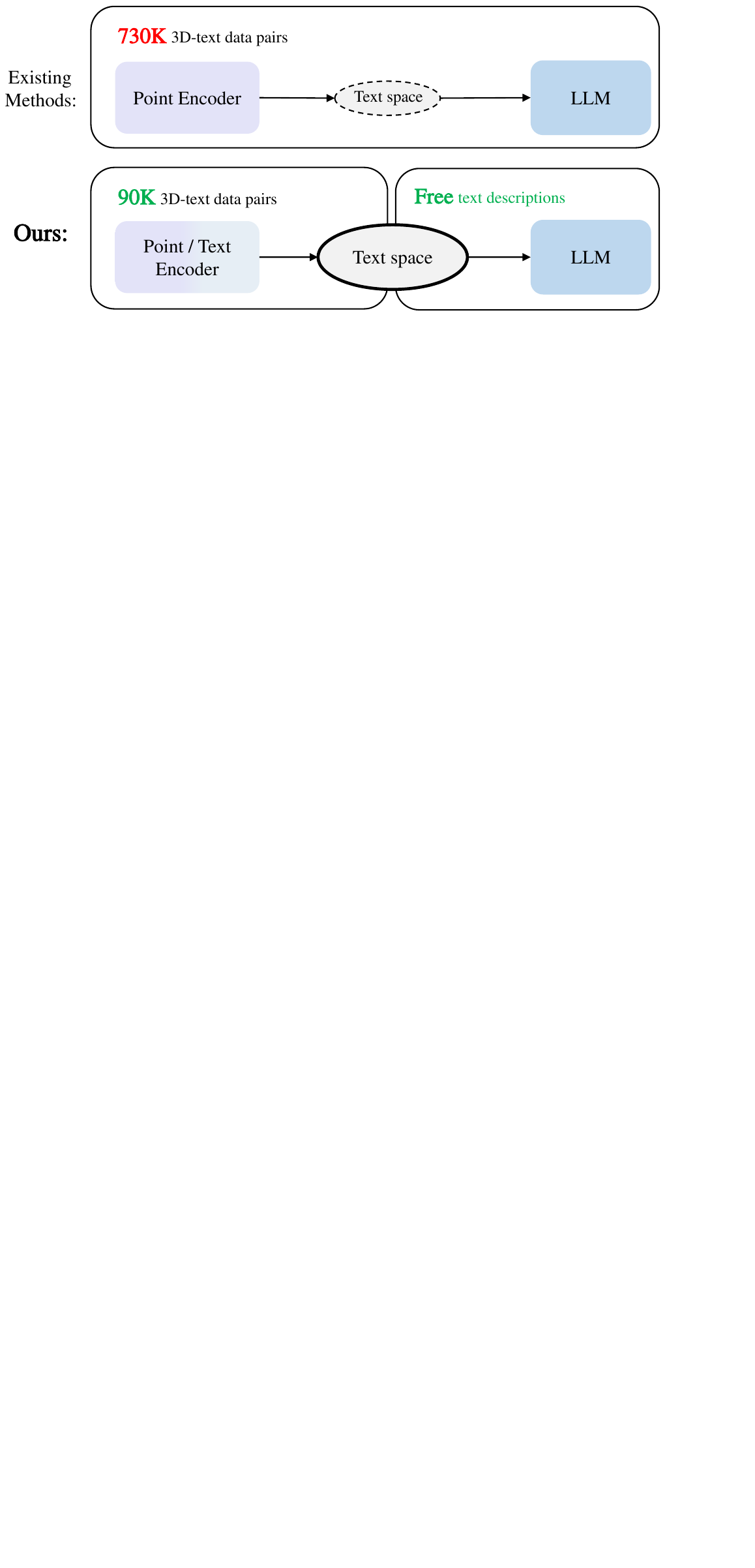}
  \caption{
     Existing methods like PointLLM use massive 3D-text data  ($\sim$730K) to enhance the point-text mapping, therefore realize point-language understanding, while we can also achieve this with only a small number of 3D data ($\sim$90K) and free-text descriptions for better point-LLM alignment.
    }
\label{fig:idea}
\end{figure}

%% file: section/03_related_work.tex
\section{Related Work}

\subsection{3D Point-Language Understanding}
To enable LLMs to understand the 3D physical world, early attempt~\cite{hong20233d} projects 3D point clouds into 2D images, relying on 2D-LLMs for comprehension. 
However, 2D-based method lose crucial 3D information, leading to issues like occlusion, ambiguity, and hallucination.
Point-Bind LLM~\cite{guo2023point} attempts to establish a 3D-2D-LLM connection, but this non-robust link leads to unstable performance.
Recently, with the availability of large-scale 3D-text data~\cite{luo2024scalable,qi2024gpt4point} and multimodal encoders, methods like PointLLM~\cite{xu2023pointllm} and ShapeLLM~\cite{qi2024shapellm} connect point encoders with LLMs and fine-tune the 3D Point Cloud-LLMs (3D-LLMs) using vast amounts of 3D-text data. 
Unfortunately, compared to images, 3D-text data remains extremely scarce (LAION-5B vs. Objaverse-1M)~\cite{schuhmann2022laion, deitke2023objaverse} and expensive, let alone the near infinite and free text data, making it challenging to build powerful 3D foundation models according to the scaling law.
Also, training 3D-LLMs is resource-intensive, often requiring 8xA100 GPUs for hundreds of hours.
Although MiniGPT-3D~\cite{tang2024minigpt} reduces training time to 26.8h on a single GPU, the 3D data bottleneck persists.
Our \ours proposes to solve this 3D data bottleneck by leveraging extensive text data to compensate for the lack of 3D data, and introducing a 3-stage training strategy for effective and efficient alignment.

\input{images/word_cloud}
\input{images/3stage}

\subsection{Multimodal Encoders in 3D-LLM}
The encoder maps raw data into a more compact embedding space, which can then be aligned with LLMs. 
To reduce the training cost, one can intuitively employ a multimodal pre-trained encoder, such as CLIP~\cite{radford2021learning}, which has been trained on text-image pairs, for aligning 2D images with LLMs. 
This makes it easier to align data from different modalities with LLMs. 
Similarly, some existing 3D-LLMs use multimodal pre-trained encoders~\cite{huang2023clip2point,xue2023ulip, qi2023contrast,gao2024sculpting, chen2024ll3da} to map point clouds into embedding space, followed by fine-tuning the  3D-LLM. 
However, even without training the encoder, constructing the 3D-LLM still requires a vast amount of point-text data~\cite{xu2023pointllm,zhou2023uni3d,qi2024shapellm,tang2024minigpt}.
We observe that existing methods underutilize the potential of the text encoder, only focusing on aligning point encoder with LLM. 
In contrast, we propose leveraging the cost-efficient text space and the text encoder to reduce the dependency on 3D data.

%% file: images/word_cloud.tex
\begin{figure}[t]
  \centering
  \begin{minipage}{0.36\linewidth}
    \centering
    \includegraphics[width=\linewidth]{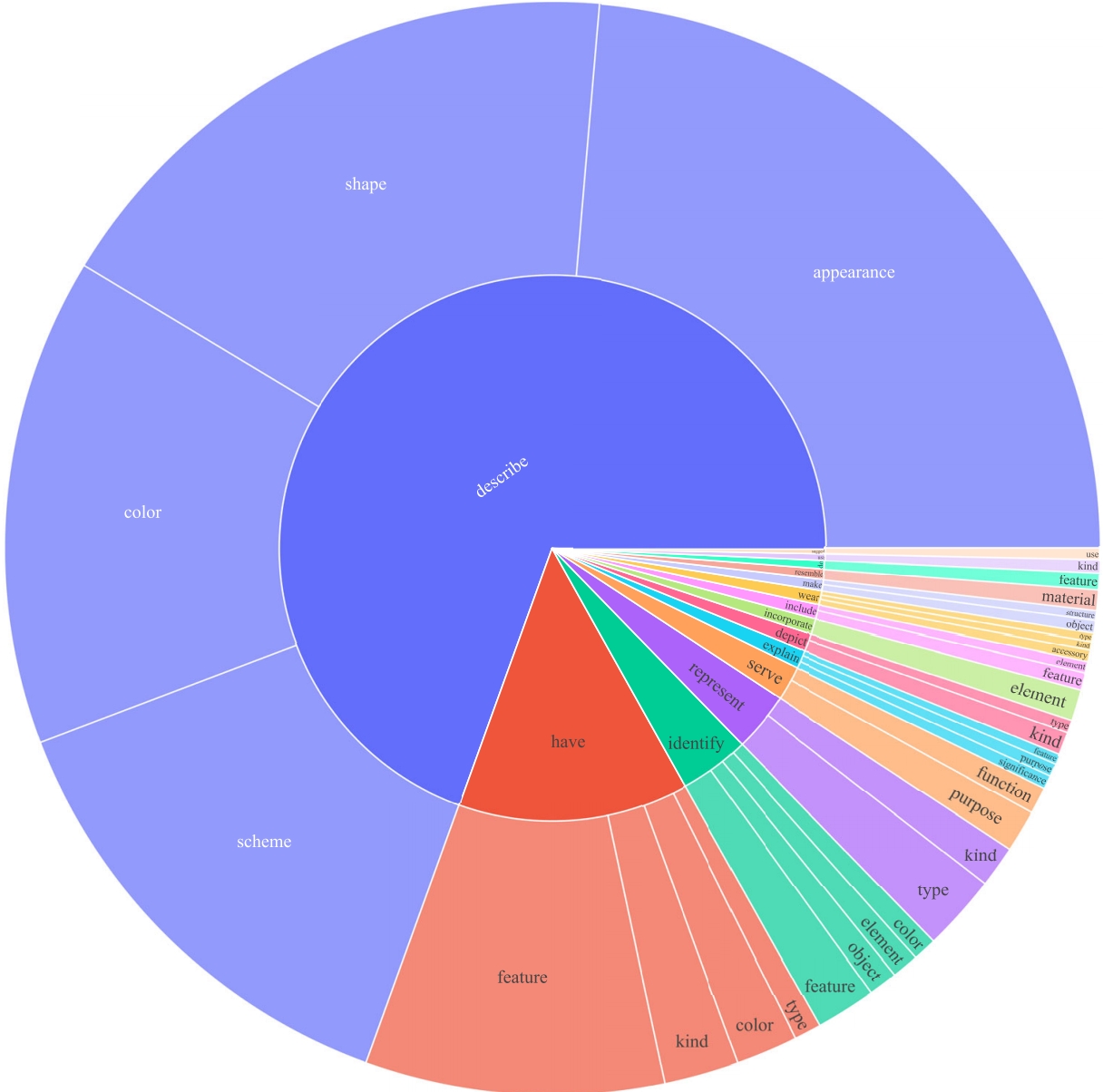}
  \end{minipage}
  \hspace{0.05cm}
  \begin{minipage}{0.61\linewidth}
    \centering
    \includegraphics[width=\linewidth]{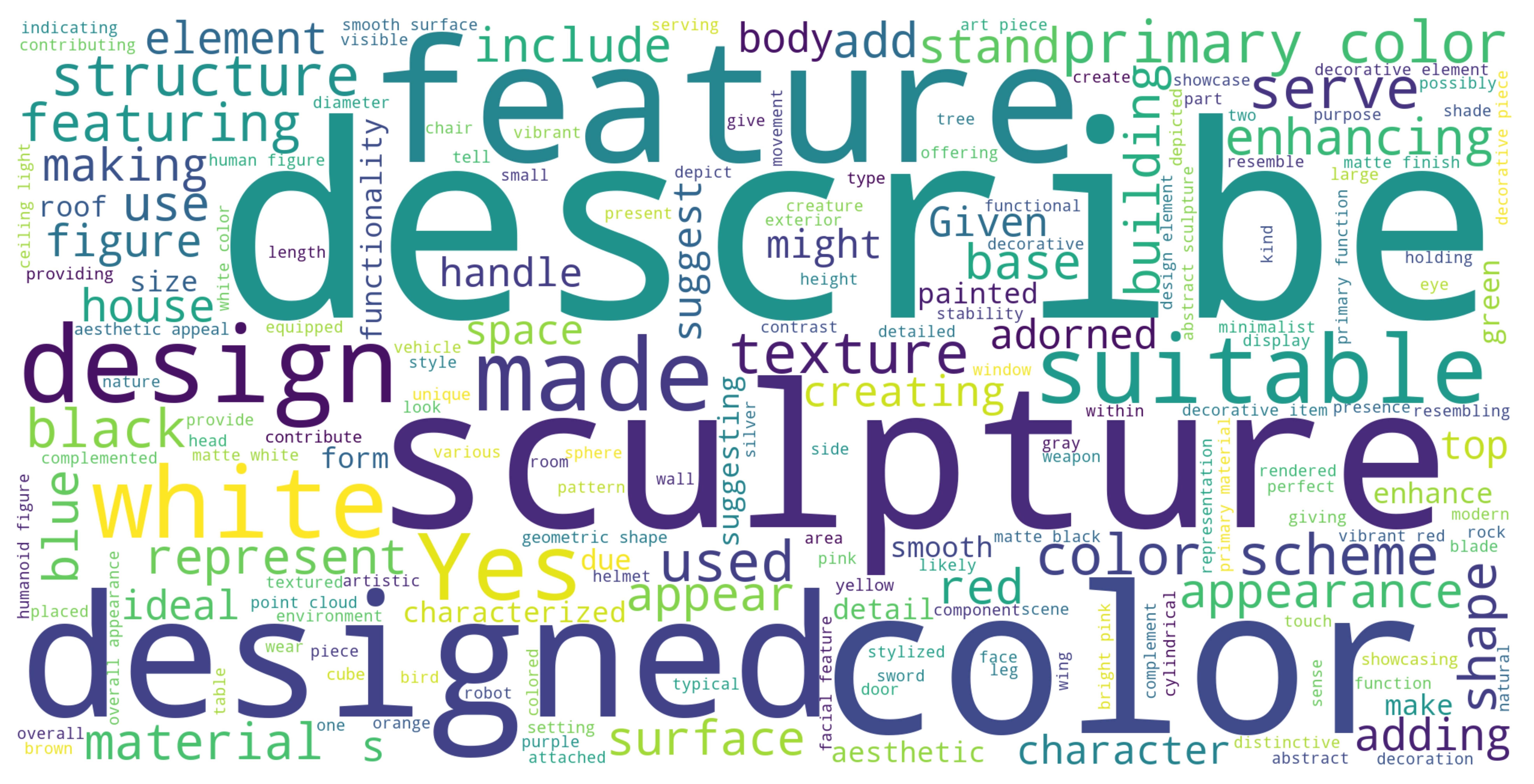}
  \end{minipage}
  \caption{T3D dataset distribution.}
    \label{fig:word_cloud}
\end{figure}

\begin{table}[t]
    \centering

    \resizebox*{\linewidth}{!}{
\begin{tabular}{l|c|l}
\toprule
Data Type                                              & Size & \multicolumn{1}{c}{Sample}                                                                                                                                                                                                   \\ \midrule
Caption                                                & 1M    & A 3D model of a friendly white dog...                                                                                                                                                                                        \\ \midrule
\begin{tabular}[c]{@{}l@{}}Brief\\ Desc.\end{tabular}  & 1M   & \begin{tabular}[c]{@{}l@{}}Q: Summarize the 3D point cloud object briefly.\\ A: The 3D object is a white dog model...\end{tabular}                                                                                           \\ \midrule
\begin{tabular}[c]{@{}l@{}}Detail\\ Desc.\end{tabular} & 1M   & \begin{tabular}[c]{@{}l@{}}Q: Offer a detailed description of this point cloud.\\ A: This 3D object depicts a white dog sitting...\end{tabular}                                                                              \\ \midrule
\begin{tabular}[c]{@{}l@{}}Single\\ Conv.\end{tabular} & 3M   & \begin{tabular}[c]{@{}l@{}}Q: What is the posture of the dog? \\ A: The dog is sitting upright. ...\end{tabular}                                                                                                          \\ \midrule
\begin{tabular}[c]{@{}l@{}}Multi\\ Conv.\end{tabular}  & 1M   & \begin{tabular}[c]{@{}l@{}}Q1: What type of animal does the 3D model represent?\\ A1: The 3D model represents a dog.\\ Q2: Can you describe the dog's posture? \\ A2: The dog is sitting upright. ...\\  ...\end{tabular} \\ \bottomrule
\end{tabular}
}
    \caption{3D object description and conversations of T3D.}
    \label{tb:t3d_example}
\end{table}

%% file: images/3stage.tex
\begin{figure*}[t]
    \centering
  \includegraphics[width=\textwidth]{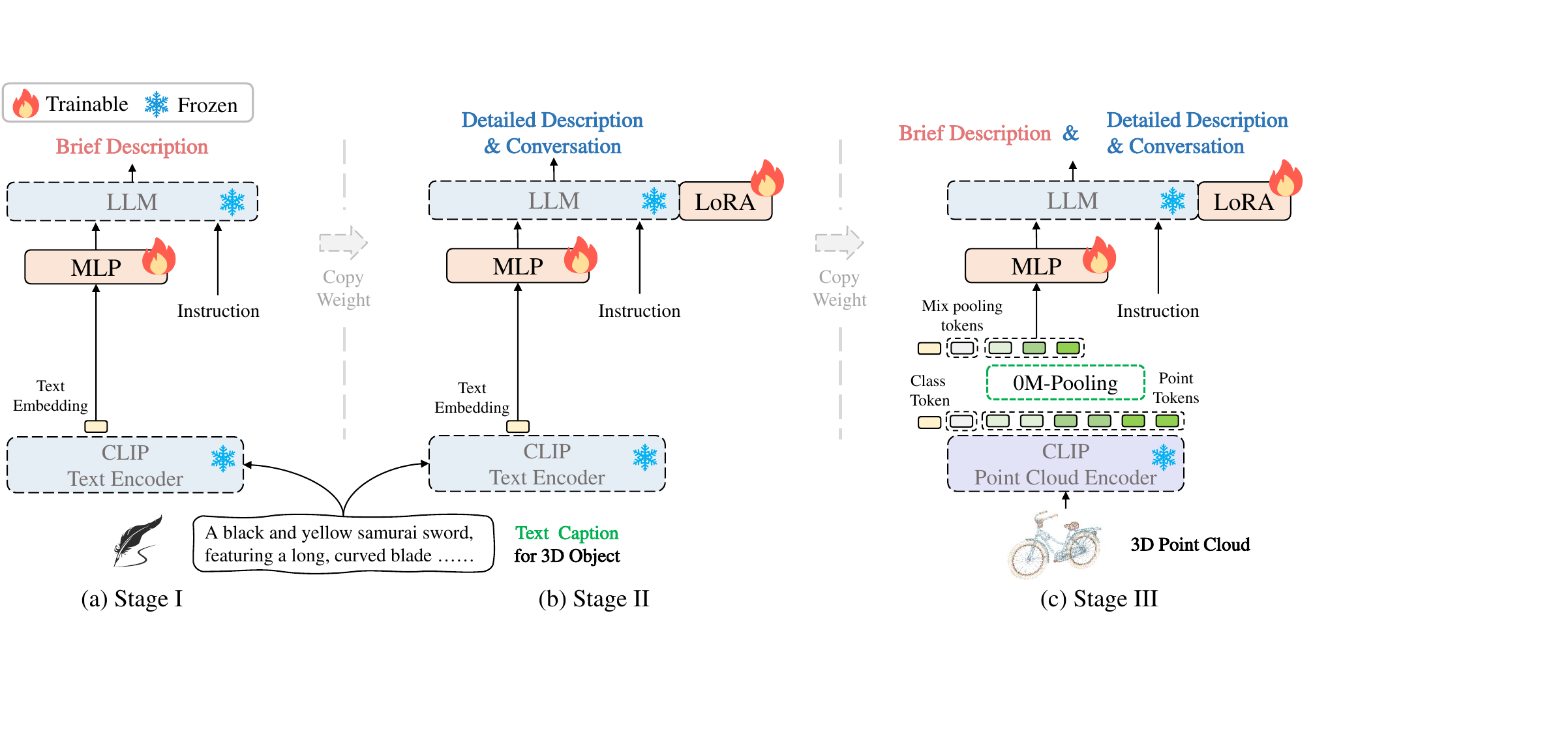}   
  \caption{
     Illustration of 3-Stage Training Strategy. We expand the text space by feeding more text data in Stage I \& II, thus reduce the demand of 3D data in Stage III. We input the text/point cloud to the encoders, then align with LLM via a MLP projector. 
     Additionally, we design a 0M-Pooling module to efficiently compress the token sequence output by point encoder.
    }
\label{3stage}
\end{figure*}

%% file: section/04_method.tex
\section{Method}
To enable LLMs to achieve robust 3D understanding with minimal 3D data, we propose using more text data to reduce reliance on 3D data. First, we generate a 6M text dataset of 3D object descriptions and conversations. Then, to better uncover connections between different modalities, we design a 3-stage training strategy. Finally, we introduce a parameter-free token pooling module to efficiently utilize information from the encoder's output token sequence. 
The details of these three parts are as follows.

\subsection{3D Object Description and Conversation Dataset}\label{sec:6mdata}
Leveraging multimodal pre-trained encoders, we propose using large amounts of text data to compensate for the lack of 3D data pairs. 
Specifically, we first align the text encoder with the LLM using extensive text data. 
Since the text encoder is already aligned with the point encoder, we then only need a small amount of 3D data for point encoder-LLM alignment.

To achieve this, we bring T3D, a 6M text dataset of 3D object descriptions and conversations. 
Fig.~\ref{fig:word_cloud} shows the verb-noun distribution and a visualized word cloud.
Instead of using the closed-source GPT-4~\cite{openai2023gpt}, we use the equally powerful open-source model Qwen2-72B-Instruct~\cite{qwen2} to construct this dataset.
We select object categories from Cap3D~\cite{luo2024scalable} and DiffuRank~\cite{luo2024view}, and we design prompts to generate 5 types of data: 1M captions, 1M brief descriptions, 1M detailed descriptions, 3M single-round conversations, and 1M multi-round conversations. 
The object descriptions help the LLMs learn rich semantic knowledge, while the conversations enable the LLMs to extract useful information from the context to improve 3D understanding.
Notably, this dataset is constructed without any manual annotation or post-processing, requiring only minimal model inference cost. 
Five types of data, totaling 6M samples in the Caption-Question-Answer format, are shown in Table~\ref{tb:t3d_example}.
During training, we input the Caption into the text encoder, pass the encoded tokens through a projector, and then input them along with the Question into the LLM, which outputs a response to calculate the loss against the Answer.
More detailed prompts and distributions are in Appendix.

\subsection{3-Stage Training Strategy}\label{sec:stage}
For better multimodal encoder-LLM alignment and minimizing the use of 3D point-text data pairs, we propose a 3-stage training strategy, as shown in Fig.~\ref{3stage}.
Our design principle is to first use a large amount of text data to align the text encoder with the LLM via a MLP projector (Stage I and II). Then, using only a small amount of 3D point-text datas, we align the point cloud encoder with the LLM via the same projector (Stage III).
Specifically, for each stage, we will introduce the pipeline, trainable layers, and data aspects as follows.

\subsubsection{Stage I}
 is shown in Fig.~\ref{3stage}(a). First, we input a   
 text caption $\boldsymbol{D}$ of a 3D object   into the pre-trained text encoder $f_{text}$, obtaining the global text embedding $\boldsymbol{C}_t$  as the encoder output. 
$\boldsymbol{C}_t$ is then passed through a learnable MLP projector $f_{proj}$ to connect with the LLM $f_{LLM}$. 
The LLM input consists of the projector output $f_{proj}(\boldsymbol{C}_t)$, and  the text tokens of an instruction prompt $\boldsymbol{I}$, such as ``What is this?''. 
Finally, the LLM outputs a brief description $\boldsymbol{R}_{brief}$ of the 3D object, which can be used to calculate the loss with the ground-truth description.
The formulas are as follows:
\begin{align}
    \boldsymbol{C}_t &= f_{text}(\boldsymbol{D}), \\
    \boldsymbol{R}_{brief} &= f_{LLM}(f_{proj}(\boldsymbol{C}_t), \text{h}(\boldsymbol{I})),
\end{align}
where $\text{h}$ is the LLM’s  tokenizer.

\textbf{\textit{Trainable Layers \& Data}:}
Note that, only the projector $f_{proj}$ is a trainable MLP, while the rest, including the text encoder $f_{text}$ and LLM $f_{LLM}$, have frozen weights.
We train the model using a large dataset of brief descriptions (1M) from our T3D dataset, as shown in  Tab.~\ref{tb:t3d_example}.

\subsubsection{Stage II}
is shown in Fig.~\ref{3stage}(b), Stage II is similar to Stage I. 
We also first input a caption of a 3D object into the text encoder $f_{text}$, then extract the global text embedding and pass it to the projector $f_{proj}$. 
The projector output, along with a complex instruction, is then fed to the LLM $f_{LLM}$. 
Finally, the LLM outputs detailed description and conversation results, which are then used to calculate the loss.

\textbf{\textit{Trainable Layers \& Data}:}
 The differences from Stage I are as follows: (1) The weights of the projector $f_{proj}$ are copied from Stage I for initialization and remain trainable.
(2) We use LoRA~\cite{hu2021lora} to train the LLM $f_{LLM}$ in this stage to achieve better multimodal alignment. 
The text encoder $f_{text}$  remains frozen.
We use only 210K detailed descriptions and conversation data for 3D objects from our T3D dataset, such as describing an object in  $\sim\!50$ words and engaging in multi-turn conversations, as shown in Tab.~\ref{tb:t3d_example}.

Notably, to enhance the perception robustness of the LLM, we add Gaussian noise to the encoder's output features to simulate the semantic discrepancies between different modalities, inspired by \citet{chen2024tomgpt}.
After two stages of pure text training, our GreenPLM acquires the ability to comprehend raw 3D point clouds by directly replacing the text encoder $f_{text}$  with a paired point encoder $f_{pc}$ from Uni3D~\cite{zhou2023uni3d}  without weight tuning.

\subsubsection{Stage III}
is shown in Fig.~\ref{3stage}(c), we use 3D point cloud as input. 
The 3D point cloud $\boldsymbol{P}$  is fed into the point cloud encoder $f_{pc}$ to output a token sequence.
Unlike previous stages that use only the global text embedding (corresponding to the class token in the point encoder) for the projector, in this stage, we extract representations from all tokens $ \boldsymbol{T}_{pc} $ to more effectively leverage information from the point  encoder.
To reduce the token sequence length for efficiency, we introduce a parameter-free token pooling module based on cross-attention, namely 0M-Pooling, which compresses the token length from 512 to 32.
The pooled point tokens $ \boldsymbol{T}_{pc}^{p} $, along with three tokens from Mix-pooling and the  class token $\boldsymbol{C}_{pc}$ , are input to the projector. 
Thus, the projector $f_{proj}$ receives 32+3+1=36 tokens.
We then feed the projector’s output, along with the instruction  $\boldsymbol{I}$, into $f_{LLM}$ to generate the predict responses  $\boldsymbol{R}_{pred}$ of descriptions or conversations. 
The responses will be used to compute loss with the ground truth.
This stage can be formulated as:
\begin{align}
    [\boldsymbol{C}_{pc}, \boldsymbol{T}_{pc}] \hspace{-0.1em} &= \hspace{-0.1em} f_{pc}(\boldsymbol{P}), 
    \ \ \boldsymbol{T}_{pc}^{p} = \text{0M-Pooling}(\boldsymbol{T}_{pc}),\\
\boldsymbol{R}_{pred } \hspace{-0.1em} &= \hspace{-0.1em} f_{LLM} \hspace{-0.2em}  \left( f_{ proj } \hspace{-0.2em} 
 \left(\boldsymbol{C}_{p c}, \operatorname{Mix}\left(\boldsymbol{T}_{pc}\right), \boldsymbol{T}_{pc}^{p}\right)\hspace{-0.2em} , \text{h}(\boldsymbol{I})\right),
\end{align}
where $\text{Mix}$ represents Mix-pooling of max, mean, and sum.

\textbf{\textit{Trainable Layers \& Data}:}
Similar to Stage II, the weights of projector $f_{proj}$ here  are copied from the previous stage and then still kept trainable. 
We continue using LoRA~\cite{hu2021lora} to train $f_{LLM}$ for efficient point-LLM alignment. 
The normalization layers and MLP in Uni3D~\cite{zhou2023uni3d} used to align the point and text encoders are fine-tuned for 0M-Pooling output, with other weights frozen.
In this stage, we train using only a small amount of 3D-text pairs (90K).

\subsubsection{Loss Function}
For all  training stages, given a pair of LLM output $\boldsymbol{R}$ and text ground truth $\boldsymbol{y}$,
GreenPLM is optimized under a causal language modeling objective~\cite{liu2018generating}:  
\begin{align}
   \mathcal{L} = \text{CrossEntropyLoss}\left(\boldsymbol{R}, \text{h}\left(\boldsymbol{y}\right) \right), 
\end{align}
where  $\text{CrossEntropyLoss}$ is the cross-entropy loss, and $\text{h}$ denotes the LLM's tokenizer.

\input{images/token}
\subsection{0M-Pooling}\label{sec:pooling}
To fully leverage the output of the point cloud encoder, we extract information from all output tokens $\boldsymbol{T}_{pc}$, not just the class token, while reducing computational load.
As shown in Fig.~\ref{fig:token}, we design a zero-parameter token pooling module based on cross-attention, namely 0M-Pooling, which compresses the 512 output tokens down to 32 tokens, without introducing any learnable parameters, defined as:
\begin{equation}
\begin{aligned}
    \boldsymbol{T}_c &= \text{FPS}(\boldsymbol{T}_{pc}), & \quad
    \boldsymbol{T}_p &= \text{KNN}(\boldsymbol{T}_c, \boldsymbol{T}_{pc}), \\
    \boldsymbol{T}_m &= \text{MaxPool}(\boldsymbol{T}_p), & \quad
    \boldsymbol{T}_{pc}^{p} &= \text{SoftMax}( ({\boldsymbol{T}_p \boldsymbol{T}_m^T})^T)\boldsymbol{T}_p,\label{equ:atten}
\end{aligned}
\end{equation}
where $\boldsymbol{T}_{pc} \in \mathbb{R}^{N \times C}$ ($N=512$) is the output point token sequence of the point cloud encoder, 
$\boldsymbol{T}_c \in \mathbb{R}^{M \times C}$  ($M=32$) is the central token gained via farthest point sampling (FPS) from $T_{pc}$,
and $\boldsymbol{T}_p  \in \mathbb{R}^{M \times K \times C}$  ($K=8$) represents the K-Nearest Neighborhood (KNN) tokens of $\boldsymbol{T}_c$ within $\boldsymbol{T}_{pc}$.
Then, we pass $\boldsymbol{T}_p$ to Max Pooling on the $K$ dimension to get $ \boldsymbol{T}_m \in \mathbb{R}^{M \times 1 \times C}$.
Finally, we use cross-attention in Equ.(\ref{equ:atten}) to aggregate information from $\boldsymbol{T}_{pc}\in \mathbb{R}^{512  \times C} \to \boldsymbol{T}_{pc}^{p}\in \mathbb{R}^{32 \times C}$.
We obtain the compressed token $ \boldsymbol{T}_{pc}^{p}$ using zero trainable parameters.
Notely, the $\boldsymbol{T}_{pc}$ input to 0M-Pooling  is from the point encoder's second-to-last layer.

%% file: images/token.tex
\begin{figure}[t]
    \centering
  \includegraphics[width=\linewidth]{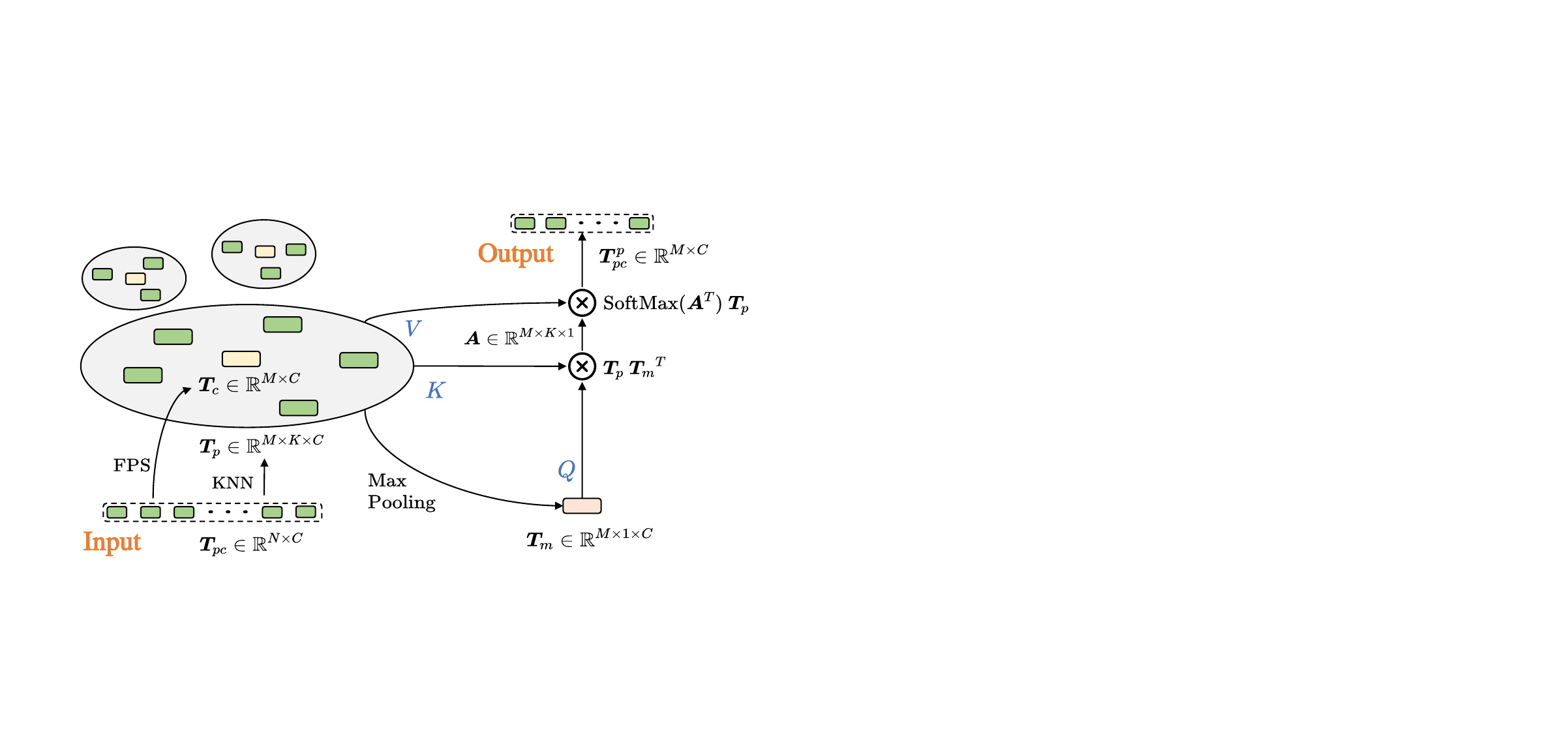}
  \caption{
     Illustration of 0M-Pooling, which compresses $N$ tokens to $M$ tokens ($M<<N$).
    }
    \label{fig:token}
\end{figure}

%% file: section/05_experiment.tex
\section{Experiment}

\textbf{Implementation details.} 
We use Phi-3~\cite{abdin2024phi} as the LLM backbone, with EVA-CLIP-E~\cite{sun2023eva} and ViT~\cite{dosovitskiy2020image} both trained by Uni3D~\cite{zhou2023uni3d} as the text encoder and point encoder, respectively. 
The point encoder outputs 512+1 tokens, each with $C=1024$. 
The MLP projector consists of two linear layers and a GeLU activation, mapping the encoder's output tokens to tokens with 3072 dimensions of  Phi-3. 
Our \ours has 63.3M trainable parameters and requires only 26.6 hours of training on a single 3090 GPU. 
Besides the standard 3-stage training of \ours, we also train \ours-0 with text-only data, utilizing only Stages I and II. 
During inference, we simply replace the text encoder in \ours-0 with the point encoder from Uni3D without weight tuning.
More detailed training settings are included in Appendix.

\paragraph{\textbf{Baselines.}}
To validate our 3D data-free capability, we compared \ours-0 with the SoTA 2D-LLMs, InstructBLIP and LLaVA, as well as the 3D-2D-LLM model Point-Bind LLM~\cite{guo2023point}.
To evaluate \ours with limited 3D data, we choose the SoTA 3D-LLMs PointLLM~\cite{xu2023pointllm} and MiniGPT-3D~\cite{tang2024minigpt}.
For fairness, we train both using the same 90K limited 3D point-text datas.

\paragraph{\textbf{Evaluation Settings.}}
An efficient and accurate model evaluation method is a shared goal in the MLLM community. 
We observe that existing evaluation approaches often rely on GPT-4 and GPT-3.5 to assess the similarity between generated results and ground truth sentences. 
While this method provides accurate evaluations, it has two major drawbacks: inconsistent API versions and high evaluation costs. 
For instance, the \textit{GPT-3.5-turbo-0613} model used in PointLLM and MiniGPT-3D is no longer maintained, making it difficult to replicate the results. 
To address these issues, we propose a new benchmark based on open-source models and introduce a new metric to evaluate data efficiency. 
Specifically, we use two prompts for the classification task: an Instruction-type (I) prompt, ``What is this?'', and a Completion-type (C) prompt, ``This is an object of.''. 
For the captioning task, we use a single prompt: ``Caption this 3D model in detail.''.
We then replace GPT-4 and GPT-3.5 with the open-source Qwen2-72B-Instruct~\cite{qwen2} (Qwen2 for short) to evaluate the model's output.
We also introduce the Accuracy-to-3D-Data Ratio (A3DR) metric to assess a model's efficiency in utilizing 3D data, defined as follows:
\begin{equation}
    \text{A3DR(Acc)} = \frac{2}{1 + \text{exp}(-\frac{\lambda\times \text{Acc}}{\text{Size} + \epsilon})} - 1,
\end{equation}
where $\text{Size}$ is the size of 3D data (K), $\text{Acc}$ is the accuracy, $\epsilon=1e-5$ prevents zero division, $\lambda=3$ adjusts discrimination.

\input{table/classification}

\input{table/captioning}

\subsection{Generative 3D Object Classification}
We validate the model's recognition ability by performing the generative 3D object classification task on the ModelNet40 dataset~\cite{wu20153d}  and Objaverse dataset~\cite{deitke2023objaverse}, using I-type and C-type prompts, with results shown in Tab.~\ref{tab:class}. 
For close-set zero-shot classification on ModelNet40, we let  Qwen2 select the closest matching category in the 40 classes as the model's output. 
For open-vocabulary classification on Objaverse, we use Qwen2 to evaluate if the model's output describes the  category of ground truth sentence.

As shown in Tab.~\ref{tab:class}, our \ours-0 achieves an average classification accuracy (AvgAcc) of 54.57\% without using any 3D data, outperforming all 2D-based models. 
It surpasses LLaVA-1.5-13B by +21.95 and Point-Bind LLM by +27.89 in AvgAcc. 
Remarkably, our model also exceeds GPT4Point (660K), which is trained with 660K 3D data, by +20.08 and performs on par with PointLLM-7B (730K).
With only a small amount of 3D data (90K), \ours achieves an average accuracy of 60.08\%, surpassing PointLLM and MiniGPT-3D by +10.95 and +11.06 in AvgAcc, respectively. 
\ours even outperforms PointLLM-13B (730K) while using a smaller LLM, and obtains results comparable to SoTA model MiniGPT-3D (730K). 
Additionally, \ours (90K) outperforms MiniGPT-3D (90K) and MiniGPT-3D (730K) on the A3DR (average accuracy) by +8.9\% and 63.1\%, respectively.
These results demonstrate the high 3D data-efficiency of our model.

\input{table/compare_demo_one}

\input{table/ablation_stage_token}

\subsection{3D Object Captioning}
We evaluate the ability to understand 3D context through a 3D object captioning task, as shown in Tab.~\ref{tab:caption}.
Following previous works~\cite{xu2023pointllm,tang2024minigpt}, we assess the similarity between the model's response and the ground truth caption using an LLM, and also evaluate embedding similarity using Sentence-BERT~\cite{reimers2019sentence} (S-BERT) and SimCSE~\cite{gao2021simcse}.

It is evident that all models without 3D data underperform compared to those trained with 3D data, as they lose significant 3D information.
However, our \ours-0 can still outperforms Point-Bind LLM and achieves comparable results to powerful 2D-LLMs by a large margin.
When using a small amount of 3D data (90K), our Qwen2 score surpasses MiniGPT-3D (90K) by +7.50, with S-BERT and SimCSE scores also exceeding by +2.72 and +1.61, respectively. 
Similarly, \ours (90K) achieves a Qwen2 score higher than PointLLM-13B (730K) by +2.15, with S-BERT and SimCSE scores comparable to MiniGPT-3D (730K) while using only 12\% of 3D data.
These results again demonstrate \ours's ability to efficiently extract 3D information from even small amounts of 3D data or purely text data.

\input{images/ablation_0m_pooling_T3D}
\input{images/stage_1_2_data}

\input{images/std_and_daily_data}

\subsection{Qualitative Results}
Fig.~\ref{fig:conv_demo} and Tab.~\ref{tab:demo} present the qualitative results.
As shown in Fig.~\ref{fig:conv_demo}, whether trained on text-only or with minimal 3D data, \ours provides accurate, context-aware responses in multi-turn conversations. 
Tab.~\ref{tab:demo} shows that our \ours-0 effectively identifies objects and understands details like color and components with text-only data.
2D-based methods like Instruct-BLIP (Ins-BLIP) and GPT-4o mini lose 3D information, suffering from occlusion, ambiguity and severe hallucinations.
Point-Bind LLM (P-B LLM) lacks accurate 3D perception due to its non-robust 3D-2D-LLM connection.
While using few 3D data (90K), \ours offers significantly more detailed descriptions and better captures local details in point clouds, such as guitar strings, compared to PointLLM.

\subsection{Ablation Study}
We conduct ablation experiments on the generative 3D object classification task and report the average accuracy.

\paragraph{\textbf{Training stages.}}
As shown in Tab.~\ref{tb:combined_ablation}, removing any stage reduces performance, with the biggest drop when Stage I is removed.
This is because Stage I trains the MLP projector to align the encoder with the LLM. 
Comparing rows \#4 and \#7, we observe that Stage II helps the LLM better align with the semantic space. 
The results of rows \#6 and \#7 indicate that Stage III injects 3D information into the LLM, significantly enhancing the model's 3D understanding.

\paragraph{\textbf{0M-Pooling.}}
As shown in Fig.~\ref{fig:ablation_OM_pooling}, when we replace 0M-Pooling with Max Pooling or Mean Pooling, the accuracy drops by 1.96 and 1.58, respectively, even though the learnable parameters remain zero. 
This demonstrates that our 0M-Pooling module effectively and efficiently captures point cloud information from the token sequence, enhancing \ours's 3D understanding ability.

\paragraph{\textbf{T3D dataset.}}
To test the impact of captions in our T3D dataset, which serve as input to the text encoder, we replace captions with low-information sentences in Stage I, and generate a 1M daily conversation dataset (example in Fig.~\ref{fig:daily_text}).
Using daily conversation data causes a significant performance drop in Fig.~\ref{fig:ablation_T3D}, indicating that captions provide more effective semantic information for the model.
Moreover, we assess the impact of text data size in Stages I and II. 
As shown in Fig.~\ref{fig:stage_1_2-data}, with more text data, the model learns from a larger text space, leading to a stronger point-text-LLM connection.
This confirms the effectiveness of the text space, reducing the need for 3D data and addressing the 3DEPL task.

\paragraph{\textbf{Token Fusion before MLP projector.}}
In Stage III, the tokens input into the MLP projector consist of three parts: the Class token, the Mix-Pooled token, and the 0M-Pooled token. 
We conduct ablation experiments on these three tokens, as shown in Tab.~\ref{tb:combined_ablation}. 
The results demonstrate that both Mix-Pooling and 0M-Pooling enhance the model's ability to extract information from the token sequence.

\paragraph{\textbf{Noise level in Stage I \& II.}}
Adding Gaussian noise to the token sequence output by the text encoder forces the LLM to learn useful information from noisy data, thereby improving the model's robustness. 
As shown in Fig.~\ref{fig:noise_std}, we experiment with different noise levels. 
As the standard deviation (std) of the noise increases from 0 to 0.06, \ours's accuracy initially increases and then decreases, reaching its peak at 0.05. 
The results demonstrate that appropriately adding noise can enhance the model's ability to extract cross-modal information, therefore improving its 3D understanding.

%% file: table/classification.tex
\begin{table*}[t]
    \centering

    \resizebox*{\linewidth}{!}{

\begin{tabular}{lcrcccccccc}
\toprule
                        &                             &                                                                       &                         &                                                                                    & \multicolumn{2}{c}{ModelNet40}  & \multicolumn{2}{c}{Objaverse}   &                           &                                                                         \\ \addlinespace[2pt] \cline{6-9} \addlinespace[2pt]
\multirow{-2}{*}{Model} & \multirow{-2}{*}{Reference} & \multirow{-2}{*}{\begin{tabular}[c]{@{}r@{}}LLM \\ Size\end{tabular}} & \multirow{-2}{*}{\begin{tabular}[c]{@{}c@{}}3D Data \\ Size\end{tabular}}& \multirow{-2}{*}{Input}  & (I)            & (C)            & (I)            & (C)            & \multirow{-2}{*}{Average} & \multirow{-2}{*}{\begin{tabular}[c]{@{}c@{}}A3DR \\ (Avg)\end{tabular}} \\ \midrule
\multicolumn{11}{c}{  {\color[HTML]{009901}   \textit{\textbf{Text-only Data}}} \textit{in Training}}                                                                                                                                                                                                                                                                                                                                                             \\ \midrule
InstructBLIP-7B~\cite{dai2024instructblip}          & NIPS23                  & 7B                                                                    & 0K   & Single-Img.                                                                                         &         17.67       &     22.81           &   21.50            &  26.00           &             22.00              & 1.000                                                                        \\
InstructBLIP-13B~\cite{dai2024instructblip}         & NIPS23                  & 13B                                                                   & 0K  & Single-Img.                                                                                         &     21.56           &      21.92          &        21.50        &  21.50              &       21.62                 &  1.000                                                                       \\
LLaVA-1.5-7B~\cite{liu2024improved}             &    CVPR24                        & 7B                                                                   & 0K   & Single-Img.                                                                                          & 27.11          & 21.68          & 37.50          & 30.00          & 29.07                     &  1.000                                                                       \\
LLaVA-1.5-13B~\cite{liu2024improved}            &      CVPR24                        & 13B                                                                  & 0K   & Single-Img.                                                                                          & 27.71          & 27.76          & 39.50    & 35.50    & 32.62              &  1.000                                                                       \\
GPT-4o mini~\cite{GPT_4o_mini}              &  OpenAI                          & -                                                                  & 0K     & Single-Img.                                                                                          & 22.00          & 23.10          & 39.00          & 35.00          & 29.78                     &  1.000                                                                       \\   \addlinespace[2pt] 
    \cline{5-11}  
    \addlinespace[2pt] 
Point-Bind LLM~\cite{guo2023point}          & arXiv23                  & 7B                                                                    & 0K & Point Cloud                                                                                           & {\ul {46.60}}    & {\ul {45.02}}    & {\ul {7.50 }}          & {\ul {7.58}}           & {\ul {26.68}}                     &                                                                      1.000   \\ \addlinespace[2pt]
\rowcolor[HTML]{EFEFEF} 
\textbf{GreenPLM-0 (Ours)}             & -                           & 3.8B                                                            & 0K    & Point Cloud                                                                                         & \textbf{62.60}   {\color[HTML]{109AFD} \textbf{(+16.00)}}& \textbf{62.68}  {\color[HTML]{109AFD} \textbf{(+17.66)}} & \textbf{48.00}  {\color[HTML]{109AFD} \textbf{(+40.50)}} & \textbf{45.00}  {\color[HTML]{109AFD} \textbf{(+37.42)}} & \textbf{54.57}   {\color[HTML]{109AFD} \textbf{(+27.89)}}          &     \textbf{1.000}                                                                    \\ \midrule
\multicolumn{11}{c}{   {\color[HTML]{01511E}     \textit{\textbf{Limited 3D Data}}} \textit{in Training}}                                                                                                                                                                                                                                                                                                                                                      \\ \midrule
PointLLM-7B~\cite{xu2023pointllm}             & ECCV24                     & 7B                                                                  & 90K   & Point Cloud                                                                                        & {\ul {45.22}}          & 39.30         & {\ul {59.00}}          & 53.00          & {\ul {49.13}}                     &                                                                     0.674    \\ \addlinespace[2pt]
MiniGPT-3D~\cite{tang2024minigpt}              & MM24                   & 2.7B                                                                & 90K     & Point Cloud                                                                                       &    43.56            &        {\ul {43.03}}        &   54.50             &  {\ul {55.00}}              &    49.02                       &                                                                       0.673  \\ \addlinespace[2pt]
\rowcolor[HTML]{EFEFEF} 
\textbf{GreenPLM (Ours)}             & -                           & 3.8B                                                              & 90K      & Point Cloud                                                                                        & \textbf{58.95}  {\color[HTML]{109AFD} \textbf{(+13.73)}} & \textbf{62.36}  {\color[HTML]{109AFD} \textbf{(+19.33)}} & \textbf{60.50}  {\color[HTML]{109AFD} \textbf{(+1.50)}} & \textbf{58.50}  {\color[HTML]{109AFD} \textbf{(+3.50)}} & \textbf{60.08}     {\color[HTML]{109AFD} \textbf{(+10.95)}}        &                                                                      \textbf{0.762}   \\ \midrule
\multicolumn{11}{c}{{\textit{Extensive 3D Data in Training}}}                                                                                                                                                                                                                                                                                                                                                       \\ \midrule
GPT4Point~\cite{qi2024gpt4point}               & CVPR24                     & 2.7B                                                                  & 660K  & Point Cloud                                                                                       & 21.39          & 21.07          & 49.00          & 46.50          & 34.49                     &                                                                      0.078   \\
PointLLM-7B~\cite{xu2023pointllm}             & ECCV24                     & 7B                                                                    & 730K   & Point Cloud                                                                                      & 51.34          & 50.36          & 62.00          & 63.00 & 56.68                     &                                                                     0.116    \\
PointLLM-13B~\cite{xu2023pointllm}            & ECCV24                     & 13B                                                                    & 730K   & Point Cloud                                                                                     & 51.70          & 52.67          & 61.50          & 63.00          & 57.22                     &                                                               0.117          \\
MiniGPT-3D~\cite{tang2024minigpt}              & MM24                   & 2.7B                                                                & 730K    & Point Cloud                                                                                       & 61.99          & 60.49          & 65.00          & 68.50          & 64.00                     &                                                                     0.131    \\ \bottomrule
\end{tabular}
} 

 \caption{Generative 3D object classification results on the ModelNet40 test split and Objaverse.
    The accuracy (\%)  under the \textbf{I}nstruction-typed (I) prompt ``What is this?'' and the \textbf{C}ompletion-type (C) prompt ``This is an object of'' are reported. 
   }
    \label{tab:class}

\end{table*}

%% file: table/captioning.tex
\begin{table*}[t]
  \centering
  \setlength\tabcolsep{15pt}

  \resizebox*{\linewidth}{!}{
\begin{tabular}{lcrccccc}
\toprule
                        &                             &                                                                       &                         &                                                                                    &                                      &                                 &                          \\
\multirow{-2}{*}{Model} & \multirow{-2}{*}{Reference} & \multirow{-2}{*}{\begin{tabular}[c]{@{}r@{}}LLM \\ Size\end{tabular}} & \multirow{-2}{*}{\begin{tabular}[c]{@{}c@{}}3D Data \\ Size\end{tabular}} & \multirow{-2}{*}{Input}  & \multirow{-2}{*}{Qwen2} & \multirow{-2}{*}{Sentence-BERT} & \multirow{-2}{*}{SimCSE} \\ \midrule
\multicolumn{8}{c}{  {\color[HTML]{009901}  \textit{\textbf{Text-only Data}}} \textit{in Training}}                                                                                                                                                                                                                                                                                          \\ \midrule
InstructBLIP-7B~\cite{dai2024instructblip}         & NIPS23                  & 7B                                                                 & 0K      & Single-Img.                                                                                         &     16.10                              &       35.79                         &        36.67                \\
InstructBLIP-13B~\cite{dai2024instructblip}         & NIPS23                 & 13B                                                                  & 0K    & Single-Img.                                                                                         &            13.79                         &       33.52                         &        35.60                \\
LLaVA-1.5-7B~\cite{liu2024improved}            &        CVPR24                      & 7B                                                                & 0K      & Single-Img.                                                                                          & 17.80                                 & 39.32                         & 41.08                  \\
LLaVA-1.5-13B~\cite{liu2024improved}              &        CVPR24                    & 13B                                                                & 0K     & Single-Img.                                                                                          & 16.00                                  & 39.64                         & 40.90                  \\
GPT-4o mini~\cite{GPT_4o_mini}              &  OpenAI                          & -                                                                  & 0K    & Single-Img.                                                                                           & 26.00                                   & 38.70                          & 39.13                 \\    \addlinespace[2pt]
    \cline{5-8}
    \addlinespace[2pt]
Point-Bind LLM~\cite{guo2023point}          & arXiv23                  & 7B                                                                  & 0K    & Point Cloud                                                                                          & {\ul {1.93}}                                 &  {\ul {27.29}}                        & {\ul {25.35}}                 \\ \addlinespace[2pt]
\rowcolor[HTML]{EFEFEF} 
\textbf{GreenPLM-0 (Ours)}              & -                           & 3.8B                                                                & 0K   & Point Cloud                                                                                           & \textbf{15.93}  {\color[HTML]{109AFD} \textbf{(+14.00)}}                               & \textbf{42.16}    {\color[HTML]{109AFD} \textbf{(+14.87)}}                      & \textbf{40.90}    {\color[HTML]{109AFD} \textbf{(+15.55)}}               \\ \midrule
\multicolumn{8}{c}{   {\color[HTML]{01511E}  \textit{\textbf{Limited 3D Data}}} \textit{in Training}}                                                                                                                                                                                                                                                                                    \\ \midrule
PointLLM-7B~\cite{xu2023pointllm}              & ECCV24                     & 7B                                                                  & 90K     & Point Cloud                                                                                      & {\ul {35.77}}                                & 46.48                       & 47.01                 \\ \addlinespace[2pt]
MiniGPT-3D~\cite{tang2024minigpt}              & MM24                   & 2.7B                                                                  & 90K   & Point Cloud                                                                                       &        35.05                              &      {\ul {46.68}}                          &        {\ul {47.75}}                  \\\addlinespace[2pt]
\rowcolor[HTML]{EFEFEF} 
\textbf{GreenPLM (Ours)}              & -                           & 3.8B                                                                 & 90K   & Point Cloud                                                                                        & \textbf{42.55}   {\color[HTML]{109AFD} \textbf{(+6.78)}}                              & \textbf{49.40 }     {\color[HTML]{109AFD} \textbf{(+2.72)}}                    & \textbf{49.36}  {\color[HTML]{109AFD} \textbf{(+1.61)}}                  \\ \midrule
\multicolumn{8}{c}{{\textit{Extensive 3D Data in Training}}}                                                                                                                                                                                                                                                                                  \\ \midrule
GPT4Point~\cite{qi2024gpt4point}               & CVPR24                     & 2.7B                                                               & 660K     & Point Cloud                                                                                       & 21.75                                & 41.10                        & 41.24                   \\
PointLLM-7B~\cite{xu2023pointllm}             & ECCV24                     & 7B                                                                  & 730K     & Point Cloud                                                                                      & 42.20                                & 48.50                        & 48.92                  \\
PointLLM-13B~\cite{xu2023pointllm}            & ECCV24                     & 13B                                                               & 730K       & Point Cloud                                                                                      & 40.40                                 & 49.07                         & 48.41                  \\
MiniGPT-3D~\cite{tang2024minigpt}              & MM24                   & 2.7B                                                                & 730K   & Point Cloud                                                                                        & 48.17                                & 49.54                         & 51.39                   \\ \bottomrule
\end{tabular}
}

     \caption{3D object captioning results on Objaverse. The results are from Qwen2 evaluation  and traditional metrics.
  %
   }
      \label{tab:caption}
\end{table*}

%% file: table/compare_demo_one.tex
\begin{table}[t]
\centering

    \resizebox*{\linewidth}{!}{
\begin{tabular}{@{}p{0.14\linewidth} p{0.86\linewidth}}
\toprule

Sample & 
  \begin{minipage}{\linewidth}
    \hspace{1.6em}
    \includegraphics[width=0.16\linewidth,height=0.12\linewidth]{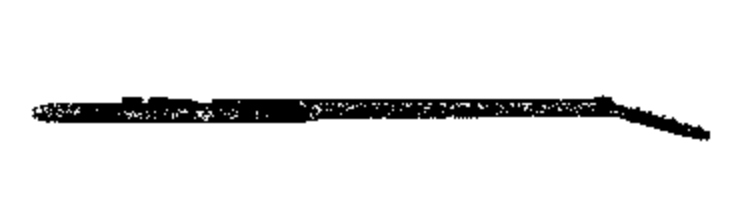}
    \hspace{3em}
    \includegraphics[width=0.16\linewidth,height=0.12\linewidth]{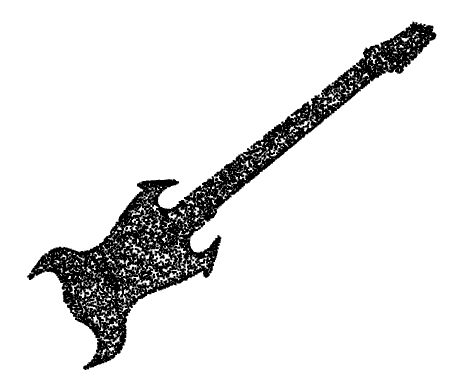}
  \end{minipage}
\\ \midrule
Prompt & What is this? \\  
Label & Guitar \\  
\midrule

Ins-BLIP & telescope \\ \addlinespace[2pt]  \cline{2-2} \addlinespace[2pt] 

GPT-4o mini  & The image appears to be a simple black outline of a tool or object, possibly resembling the shape of a knife or a similar implement. \\ \addlinespace[2pt] \cline{2-2} \addlinespace[2pt] 

P-Bind & This is a bird flying in the sky. \\ \addlinespace[2pt] \cline{2-2} \addlinespace[2pt]
 \rowcolor[HTML]{EFEFEF} 
\textbf{Ours}  ({\color[HTML]{009901} Text-only} )& The object is a Gibson Les Paul electric guitar, predominantly black with a glossy finish. It has a maple fretboard, a single-coil pickup, and a vibrato bridge. The guitar's body is adorned with the Gibson logo, and it includes a strap button for easy carrying. \\
\midrule
PointLLM & This 3D model represents a black electric guitar equipped with a distinctive headstock. \\ \addlinespace[2pt] 
\cline{2-2} \addlinespace[2pt] 
 \rowcolor[HTML]{EFEFEF} 
\textbf{Ours} ({\color[HTML]{009901} Limited 3D data}) & This is a 3D model of a cartoon-style electric guitar. The guitar is predominantly black, giving it a sleek and modern appearance. The design is simplified and stylized, typical of cartoon aesthetics, making it suitable for use in animated films, video games, or other digital media. Despite its cartoonish appearance, it retains the recognizable features of an electric guitar, such as the fretboard and strings. \\
\bottomrule

\end{tabular}
}
\caption{Qualitative comparisons. Conversation example of a guitar in ModelNet40. Our \ours generates more detailed and insightful responses compared to others.}
\label{tab:demo}
\end{table}

%% file: table/ablation_stage_token.tex
\begin{table}[t]
    \centering

    \resizebox*{\linewidth}{!}{
    \begin{tabular}{c|cccc}
    \toprule
    \#No. & Stage I & Stage II & Stage III & Acc.  \\ 
    \midrule
    1      & $\checkmark$  &          &           & 53.85 \\
    2      &         & $\checkmark$ &           & 47.03 \\
    3      &         &          & $\checkmark$ &  45.29     \\
    4      & $\checkmark$ &          &  $\checkmark$  &   58.25    \\
    5      &         & $\checkmark$ & $\checkmark$ &    42.78   \\
    \rowcolor[HTML]{EFEFEF} 
    6      & $\checkmark$ & $\checkmark$ &           & \textbf{54.57} \\
    \rowcolor[HTML]{EFEFEF} 
    7      & $\checkmark$ & $\checkmark$ & $\checkmark$ & \textbf{60.08} \\ 
    \midrule
    \#No. & Class Token & Global Tokens & Pooled Point Tokens & Acc.  \\ 
    \midrule
    8      & $\checkmark$  &          &           & 38.36 \\
    9      & $\checkmark$  & $\checkmark$ &           & 45.42 \\
    \rowcolor[HTML]{EFEFEF} 
    10     & $\checkmark$  & $\checkmark$ & $\checkmark$ & \textbf{60.08} \\
    \bottomrule
    \end{tabular}
    }
    \caption{Ablation on 3-Stage Training and Token Fusion.}
    \label{tb:combined_ablation}
    
\end{table}

%% file: images/ablation_0m_pooling_T3D.tex
\begin{figure}[t]
	\centering
	\begin{minipage}[b]{0.52\linewidth}
		\centering
		\includegraphics[width=\linewidth]{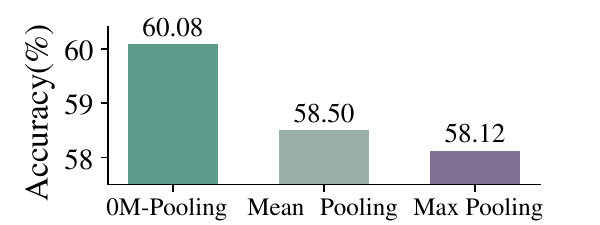}
		\captionsetup{justification=centering} 
		\caption{Ablation on 0M-Pooling.}
            \label{fig:ablation_OM_pooling}		
	\end{minipage}
	\hspace{0.1cm}
	\begin{minipage}[b]{0.38\linewidth}
		\centering
		\includegraphics[width=0.93\linewidth]{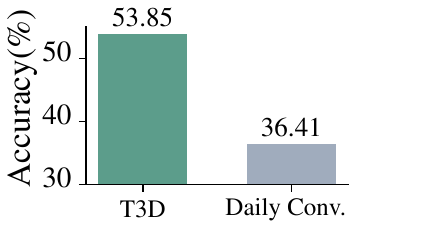}
		\captionsetup{justification=centering} 
		\caption{Ablation on T3D caption.}
            \label{fig:ablation_T3D}
	\end{minipage}
\end{figure}

%% file: images/stage_1_2_data.tex
\begin{figure}[t]
  \centering
  \begin{minipage}{0.485\linewidth}
    \centering
    \includegraphics[width=\linewidth]{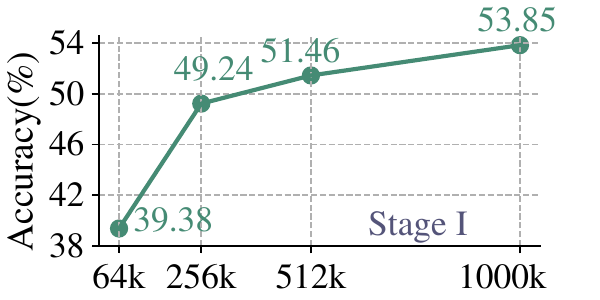}
    \centering 
  \end{minipage}
  \hspace{0.01cm}
  \begin{minipage}{0.485\linewidth}
    \centering
    \includegraphics[width=\linewidth]{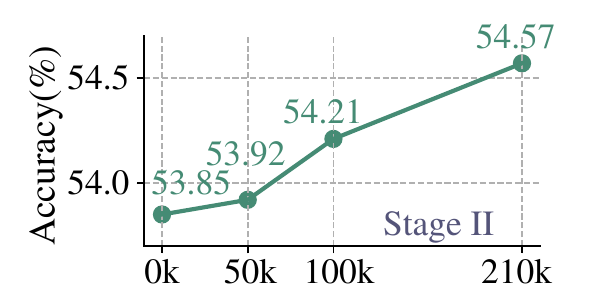}
    \centering 
  \end{minipage}

  \caption{Ablation on Text data size in Stage I \& II.
    %
    }
    \label{fig:stage_1_2-data}
    
\end{figure}

%% file: images/std_and_daily_data.tex
\begin{figure}[t]
	\centering
    \begin{minipage}{0.46\linewidth}
		\centering
		\includegraphics[width=0.93\linewidth]{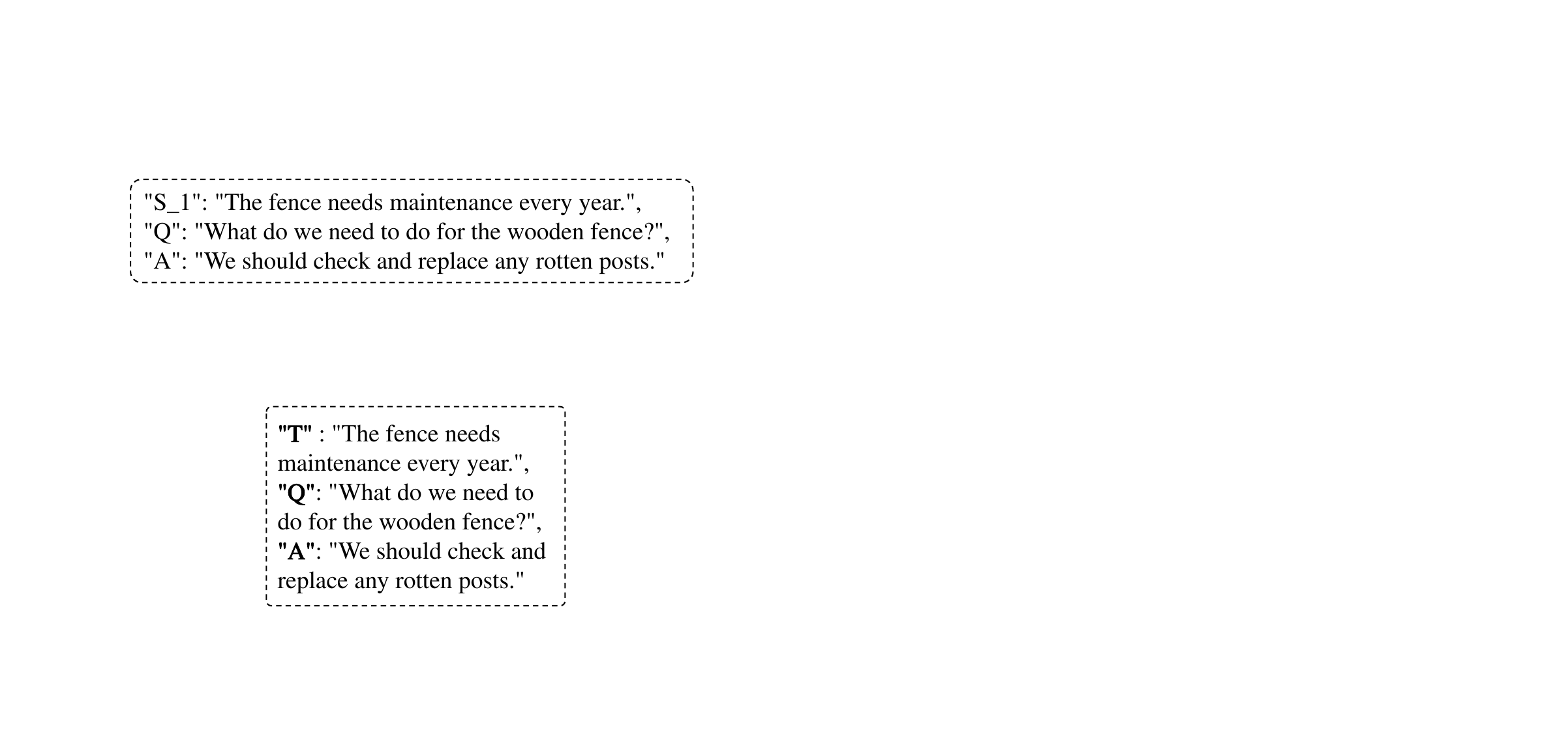}	
		\caption{Daily text data.}
            \label{fig:daily_text}
	\end{minipage}
	\begin{minipage}{0.51\linewidth}
		\centering
		\includegraphics[width=\linewidth]{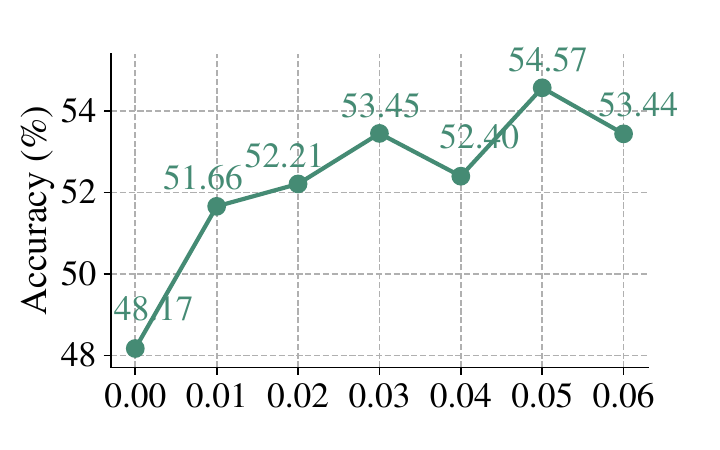}		
		\caption{Noise Std.}
            \label{fig:noise_std}
	\end{minipage}
\end{figure}

%% file: section/06_conclusion.tex
\section{Conclusion}
To enable LLMs to achieve strong 3D understanding with minimal 3D data, we introduce a new task: 3D Data-Efficient Point-Language Understanding. 
We propose \ours, which employs a 3-stage training strategy that increases text data in Stage I \& II to reduce the need for 3D data in Stage III. 
We create a 6M T3D dataset and an unified benchmark.
Results show that \ours achieves performance comparable to state-of-the-arts using only 12\% of the 3D data. 
Remarkably, our model performs well even without 3D data.

\textbf{Limitations}. 
Our approach has limitations. 
Due to time and resource constraints, we couldn't explore all text and 3D data combinations. 
We believe scaling up either could further improve performance. 
Additionally, we only test feasibility on small objects, and will explore \ours's potential for larger scenes in future work.

%% file: section/07_ack.tex
\section*{Acknowledgments}
This work was supported by the China National Natural Science Foundation No. 62202182, No. 62176101, No. 62276109,  and also supported by Guangdong Basic and Applied Basic Research Foundation 2024A1515010224, 2024A1515030017 and 2024A1515011153.

%% file: section/10_appendix.tex
\section{Appendix}

Here in the Appendix, we present the detailed distribution of the T3D dataset, along with the prompts and instructions used to create it, and provide several data examples. 
We  also showcase more visual result comparisons.
Additionally,  we provide additional ablation results and more detailed training parameters. 
Finally, we include illustrations of the model architecture used during training and inference.

\subsection{Our 6M T3D Dataset}
\subsubsection{Distributions}
We show the detailed distributions of our 6M T3D dataset in Fig.~\ref{fig:abla_word_cloud} and Fig.~\ref{fig:abla_sentence_len}.
Specifically, in Fig.~\ref{fig:abla_word_cloud}, we show word clouds for captions and responses. 
Following \citet{wang2022self}, we also present the distribution of verb-noun pairs in the dataset, highlighting its diverse attributes. 
Additionally, in Fig.~\ref{fig:abla_sentence_len}, we display the length distribution for different data types; for instance, most brief and detailed descriptions range from about 18 to 42 words.

\subsubsection{Prompts and Instructions}
Here, we show one example of the data generation pipeline using Qwen2-72B-Instruct\cite{qwen2short} in Fig.~\ref{fig:abla_Qwen_generation}, also the instruction list for description data in Fig.~\ref{fig:abla_brief_inst} and Fig.~\ref{fig:abla_detail_inst}.

\subsubsection{Data Samples}
We give several samples of four types of data from our T3D dataset, shown in Fig.~\ref{fig:abla_sample_brief}-\ref{fig:abla_sample_multi_conv}. 
Please see supplementary material for more data samples of our T3D dataset.

\subsection{Qualitative Results}
We show more qualitive results of our \ours-0 and \ours in Fig.~\ref{fig:abla_conv_text_only}  and Fig.~\ref{fig:abla_conv_stage_3}. 
Trained with only text data, our \ours-0 accurately identifies the shape, color, and usage of 3D objects. 
For example, in the bottom right of Fig.~\ref{fig:abla_conv_text_only}, the model not only correctly recognizes the shoe's category and purpose but also accurately distinguishes between the left and right shoe.
When using only a small amount of 3D data, our \ours can accurately and thoroughly describe 3D objects, as shown in the first row of Fig.~\ref{fig:abla_conv_stage_3}.

\subsection{Detailed Training Settings}
We report more detailed training settings in Tab.~\ref{tb:abla_train_setting}.

\subsection{{Ablation on the size of point encoder}}
\input{table/ablation_point_encoder_scale}
We conduct ablation experiments on the generative 3D object classification task and report the average accuracy.
We experiment with 5 different sizes of point encoders, with parameters ranging from 6.2M to 1016.5M, as shown in Tab.~\ref{tb:ablation_point_encoder_scale}. 
The results indicate that as the point encoder size increases, the model's performance first improves and then declines, achieving the best accuracy with 22.6M parameters. 
Notably, even with just 6.2M parameters, \ours still demonstrates strong 3D understanding, further proving the efficiency of our model.

\subsection{Training and Inference Architecture}
We show the difference of architectures between training and inference in Fig.~\ref{fig:abla_train_inference}.
Note that each stage of our 3-stage strategy can be used for inference. 
Specifically, if using the Stage I or Stage II for inference, simply replace the text encoder with the aligned point encoder.

\input{Appendix/image/T3D_dist}

\input{Appendix/image/T3D_len}

\input{Appendix/image/Qwen_generation_data}

\input{Appendix/table/instruct_list}

\input{Appendix/image/sample_brief}
\input{Appendix/image/sample_detail}

\input{Appendix/image/sample_single_conv}

\input{Appendix/image/sample_multi_conv}

\include{Appendix/image/conv_text_only}

\include{Appendix/image/conv_stage_3}

\input{Appendix/table/setting}

\input{Appendix/image/tain_inference}

%% file: table/ablation_point_encoder_scale.tex
\begin{table}[h]
    \centering
    \small

    \resizebox*{0.8\linewidth}{!}{
\begin{tabular}{c|rc}
\toprule
Point Encoder &Param Size  & Avg. Acc.           \\ \midrule
Giant        &   1016.5M    & 48.36          \\
Large        &  306.7M     & 54.50          \\
Base         & 88.4M      & 52.74          \\
\rowcolor[HTML]{EFEFEF} 
Small       &  22.6M      & \textbf{54.57} \\
Tiny        &  6.2M      & 54.34          \\ \bottomrule
\end{tabular}
}

\caption{Ablation on point encoder size.}
\label{tb:ablation_point_encoder_scale}
   
\end{table}

%% file: Appendix/image/T3D_dist.tex
\begin{figure*}[t]
  \centering
  \subfigure[Text caption for 3D object.]{
  \begin{minipage}{0.48\linewidth}
    \centering
    \includegraphics[width=\linewidth]{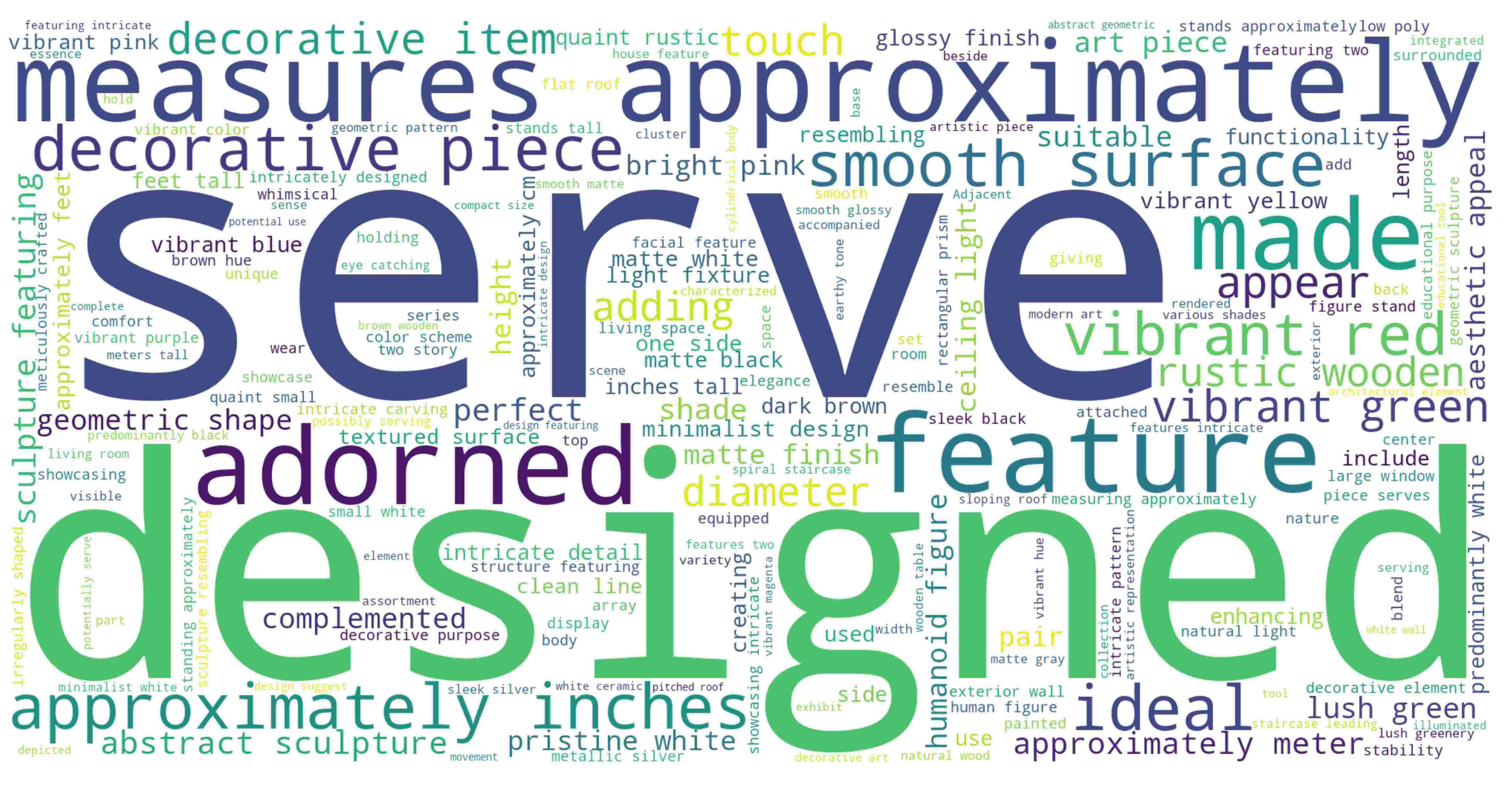}

  \end{minipage}
  }

  \vspace{0.5em} 
\centering
\subfigure[Brief description response.]{
  \begin{minipage}[b]{0.48\linewidth}
    \centering
    \includegraphics[width=\linewidth]{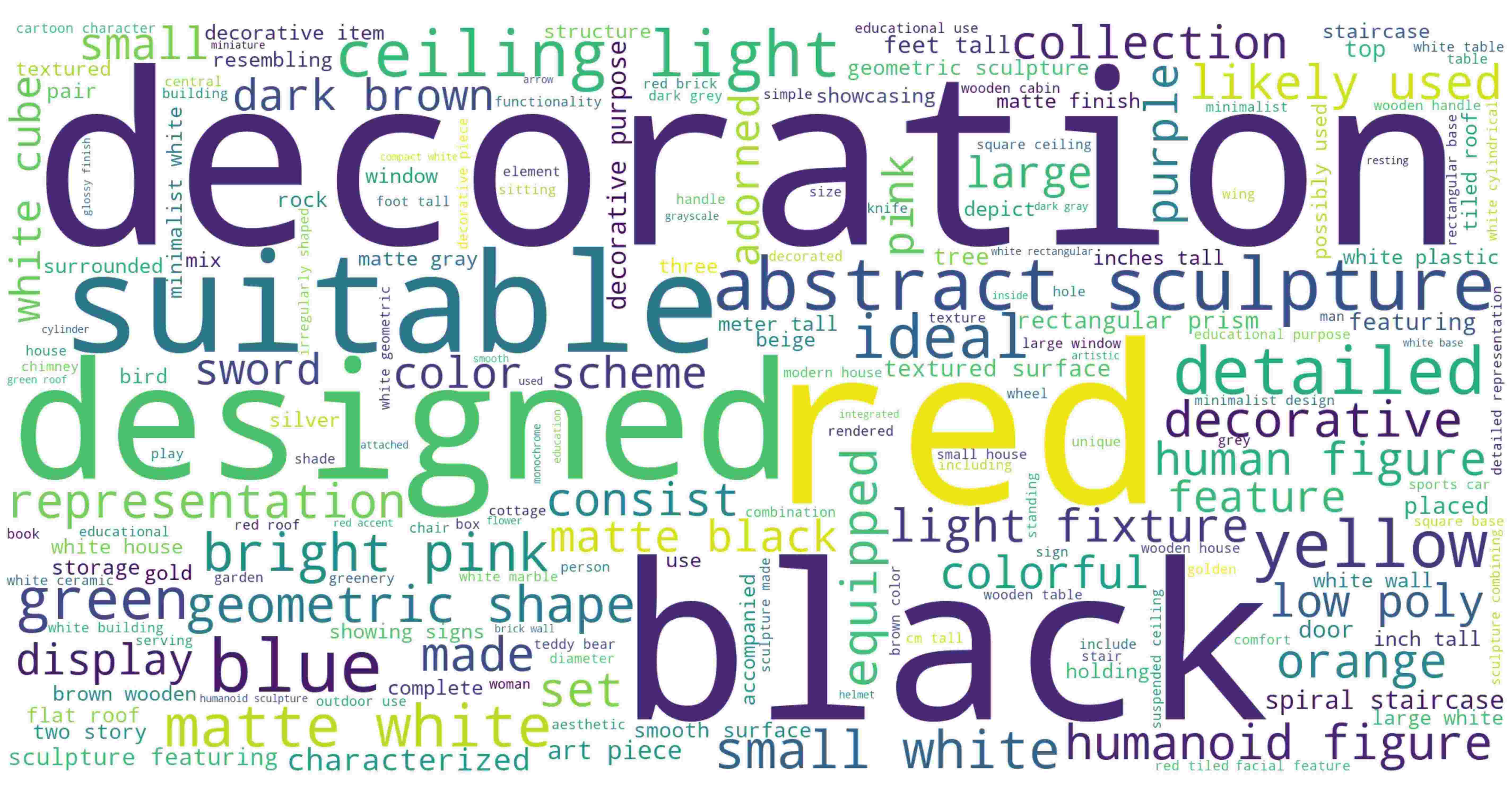}
  \end{minipage}
  }
  \subfigure[Detailed description response.]{
  \begin{minipage}[b]{0.48\linewidth}
    \centering
    \includegraphics[width=\linewidth]{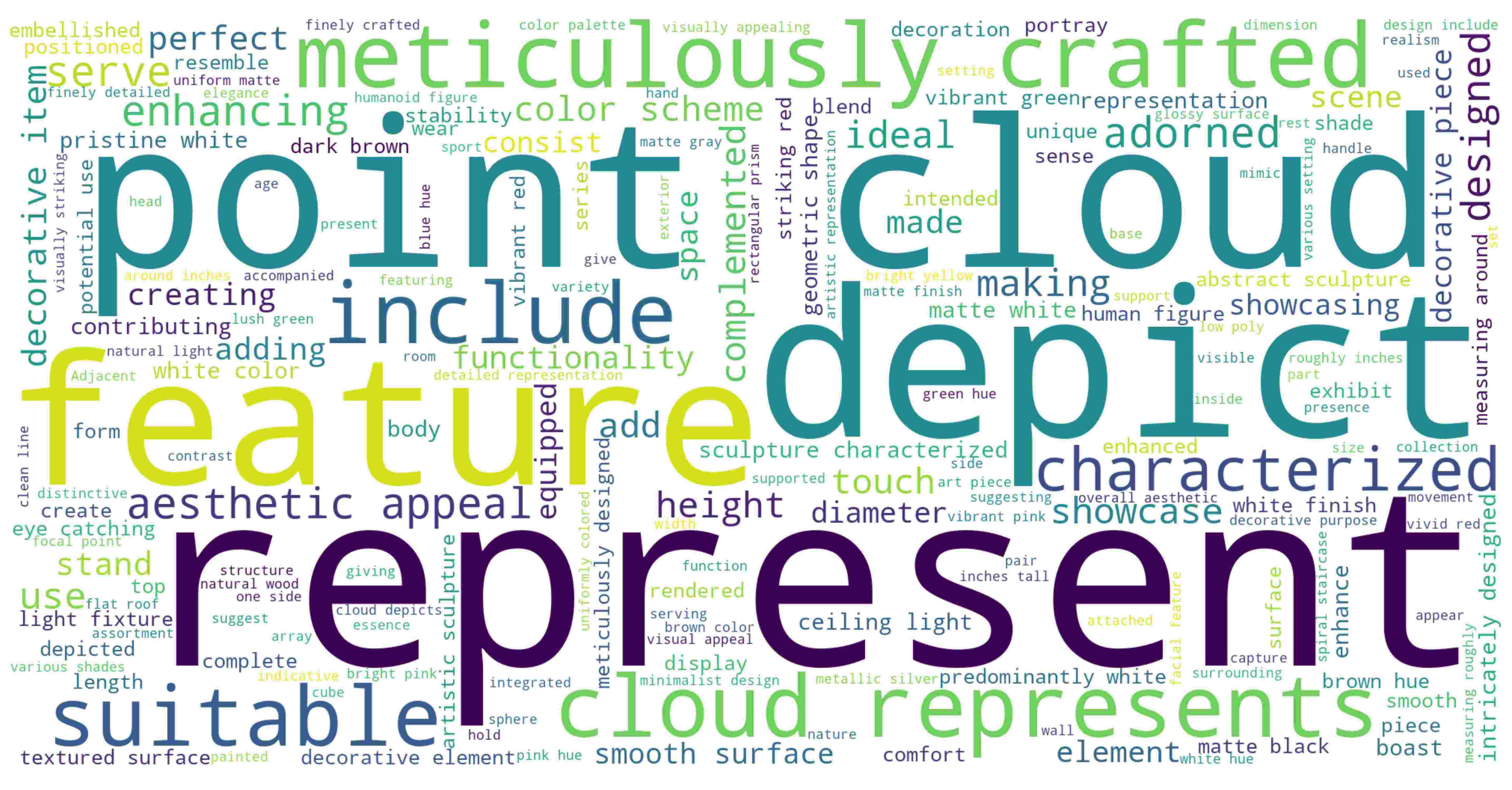}
  \end{minipage}
  }

  \vspace{1em} 
  \centering
  \subfigure[Single-round instruction.]{
  \begin{minipage}[b]{0.30\linewidth}
    \centering
    \includegraphics[width=\linewidth]{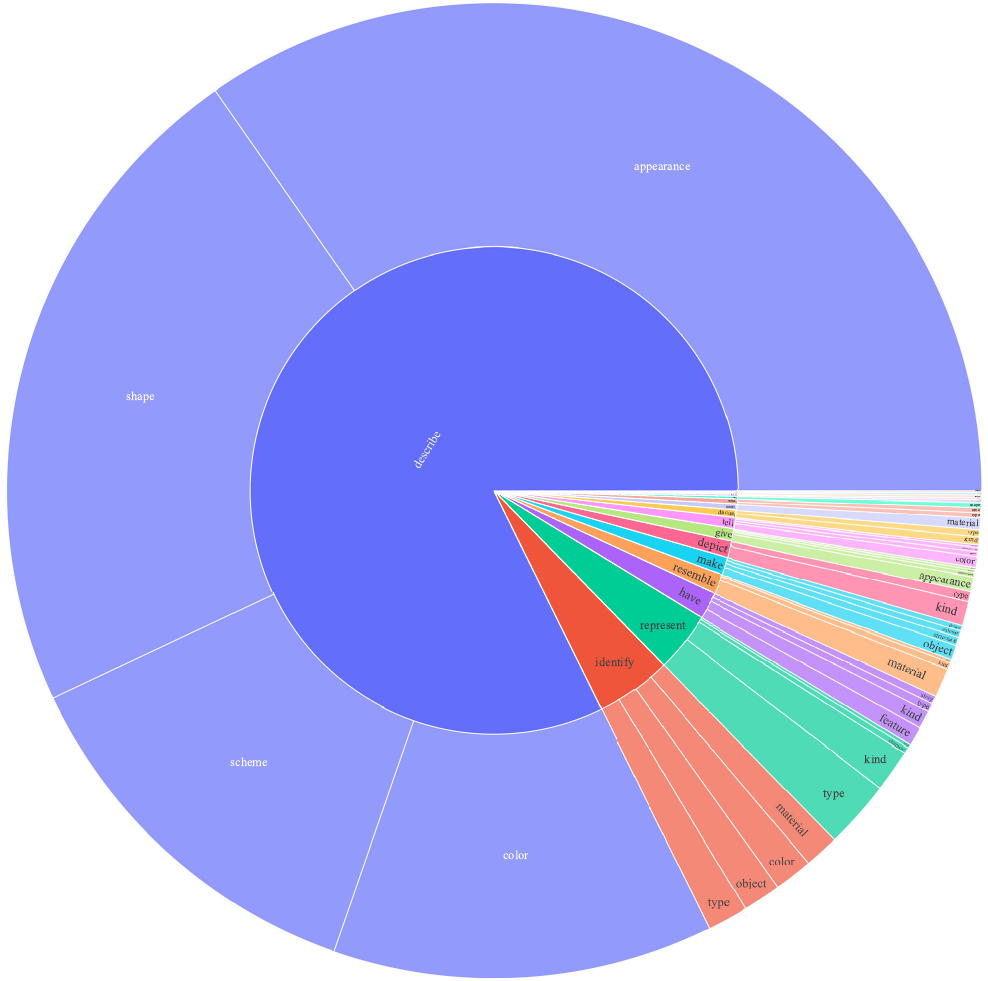}
  \end{minipage}
  }
  \hspace{0.1cm}
  \subfigure[Single-round response.]{
  \begin{minipage}[b]{0.48\linewidth}
    \centering
    \includegraphics[width=\linewidth]{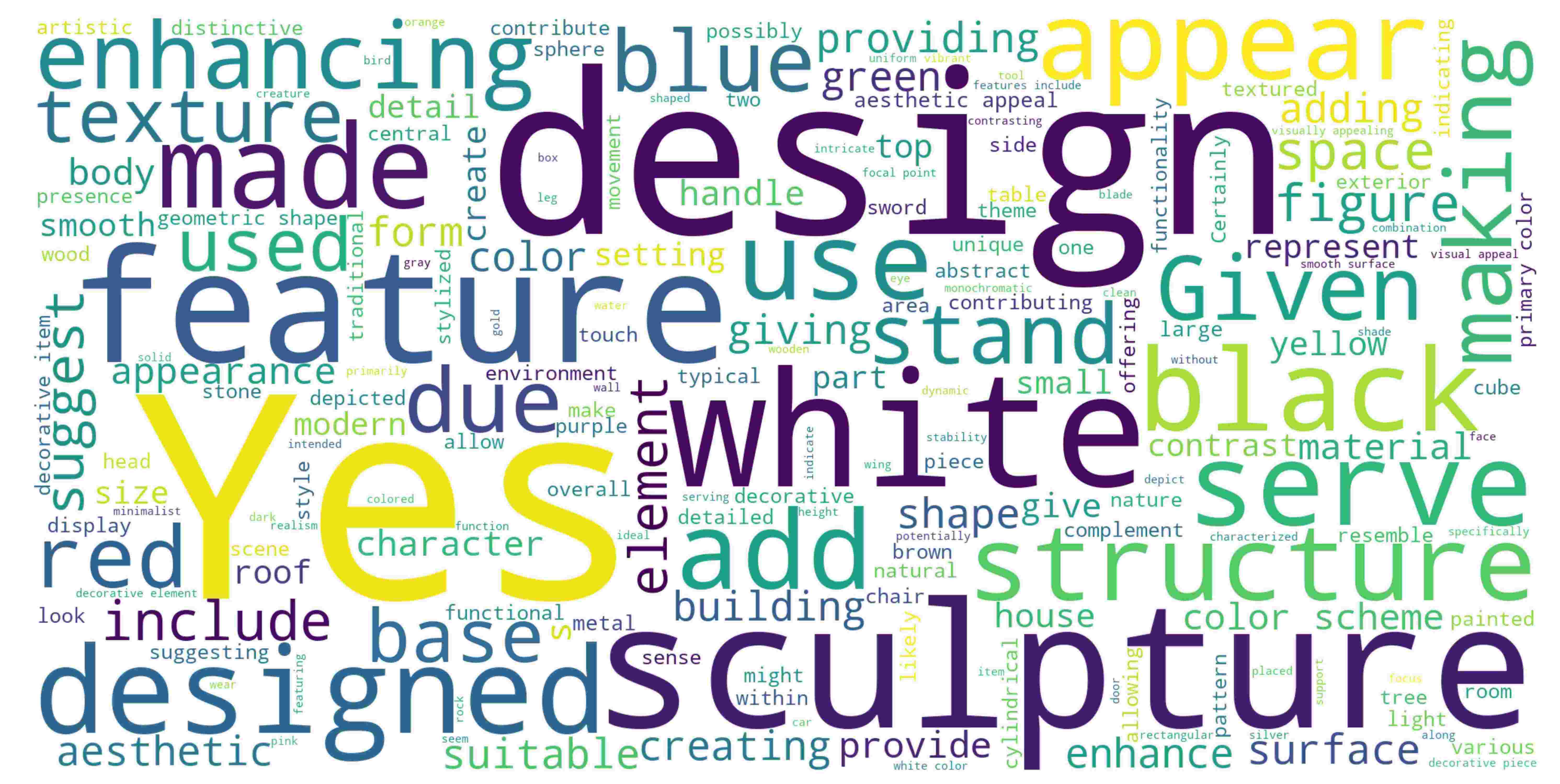}
  \end{minipage}
  }

  \vspace{1em} 
    \centering
    \subfigure[Multi-round instruction.]{
  \begin{minipage}[b]{0.30\linewidth}
    \centering
    \includegraphics[width=\linewidth]{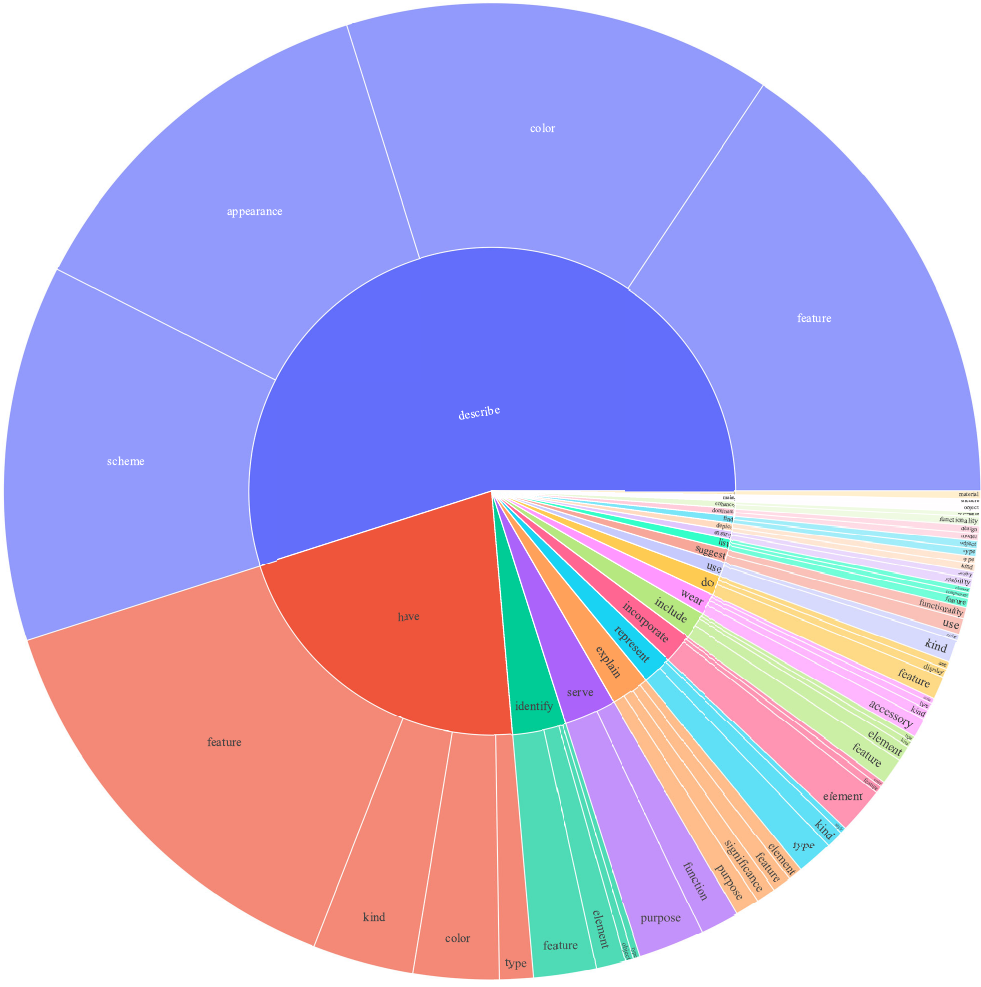}
  \end{minipage}
  }
   \hspace{0.1cm}
   \subfigure[Multi-round response.]{
  \begin{minipage}[b]{0.48\linewidth}
    \centering
    \includegraphics[width=\linewidth]{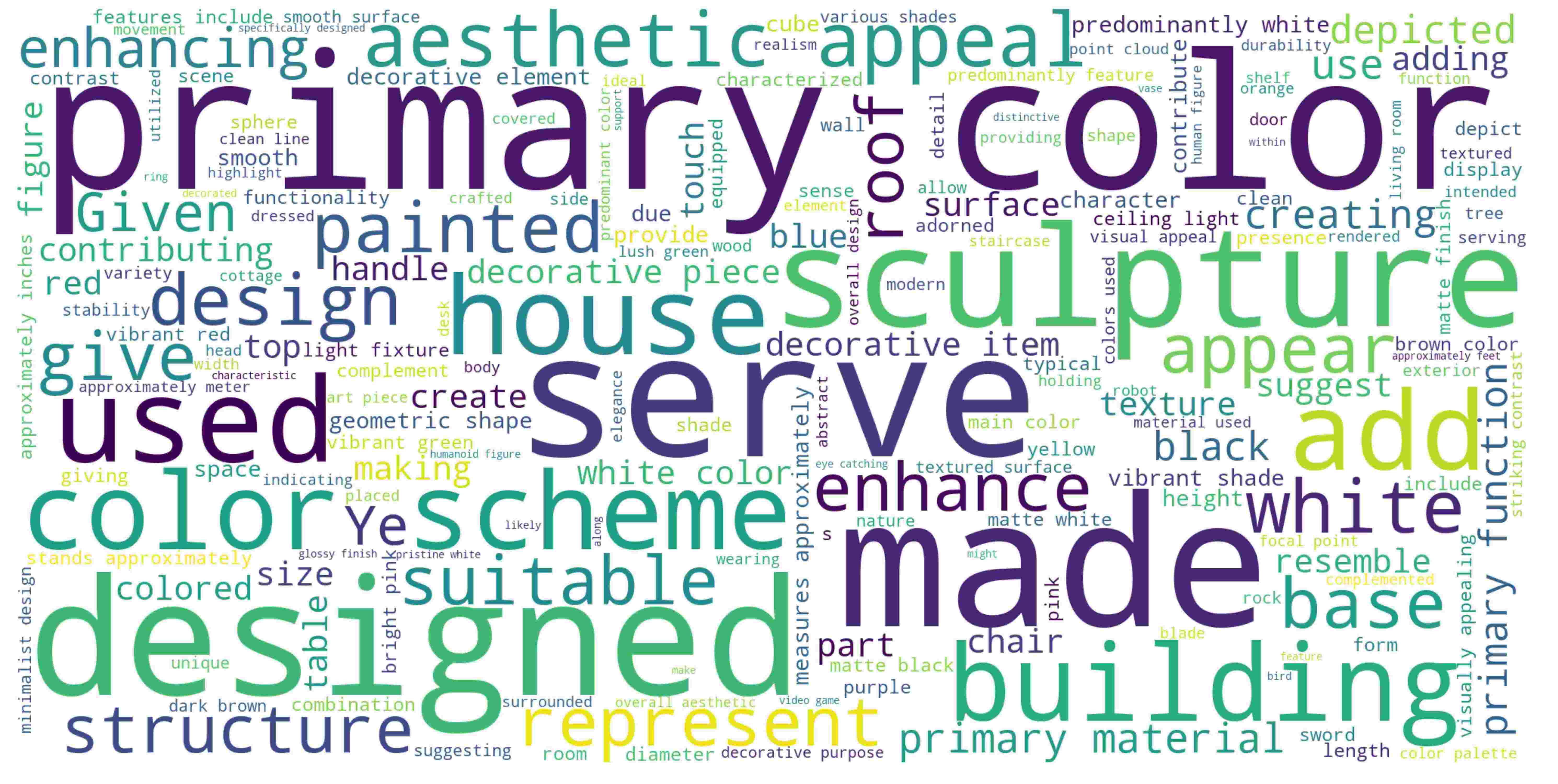}
  \end{minipage}
  }

  \caption{\textbf{Word clouds and verb-noun pair distributions of all types of data in our T3D dataset.}}
  \label{fig:abla_word_cloud}

\end{figure*}

%% file: Appendix/image/T3D_len.tex
\begin{figure*}[t]

  \vspace{0.25em} 
\centering
\subfigure[Brief description instruction.]{
  \begin{minipage}[b]{0.48\linewidth}
    \centering
    \includegraphics[width=\linewidth]{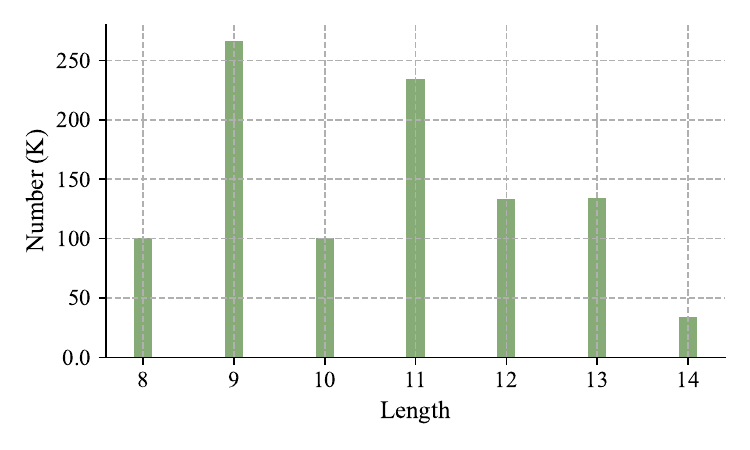}
  \end{minipage}
  }
  \subfigure[Brief description response.]{
  \begin{minipage}[b]{0.48\linewidth}
    \centering
    \includegraphics[width=\linewidth]{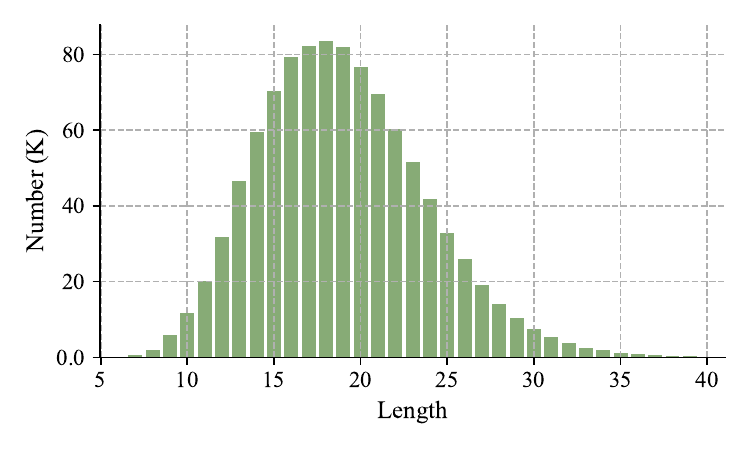}
  \end{minipage}
  }

  \vspace{0.25em} 
  \centering
  \subfigure[Detailed description instruction.]{
  \begin{minipage}[b]{0.48\linewidth}
    \centering
    \includegraphics[width=\linewidth]{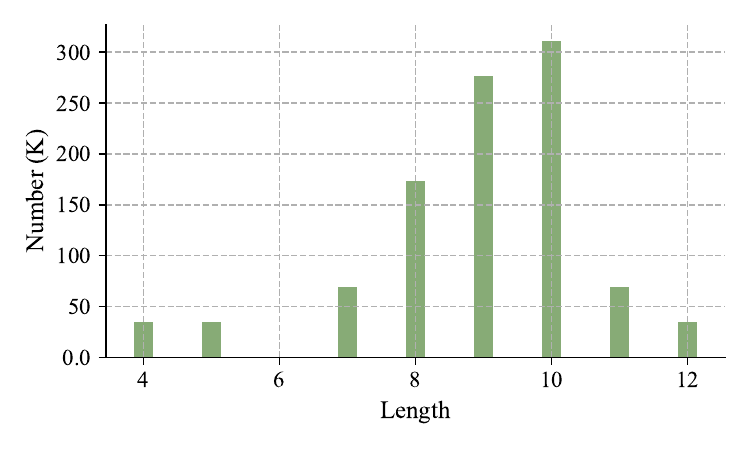}
  \end{minipage}
  }
  \hspace{0.05cm}
  \subfigure[Detailed description response.]{
  \begin{minipage}[b]{0.48\linewidth}
    \centering
    \includegraphics[width=\linewidth]{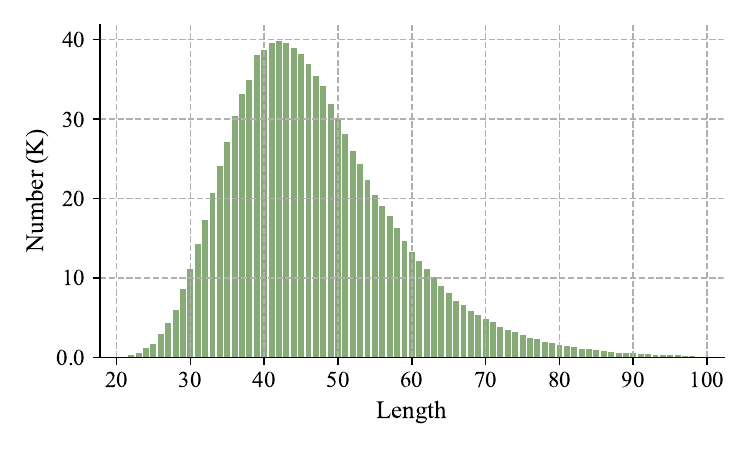}
  \end{minipage}
  }

  \vspace{0.25em} 
    \centering
    \subfigure[Single-round instruction.]{
  \begin{minipage}[b]{0.48\linewidth}
    \centering
    \includegraphics[width=\linewidth]{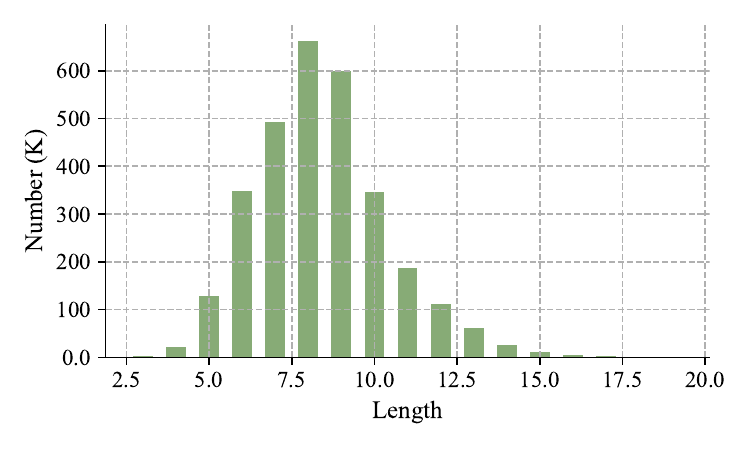}
  \end{minipage}
  }
   \hspace{0.05cm}
   \subfigure[Single-round response.]{
  \begin{minipage}[b]{0.48\linewidth}
    \centering
    \includegraphics[width=\linewidth]{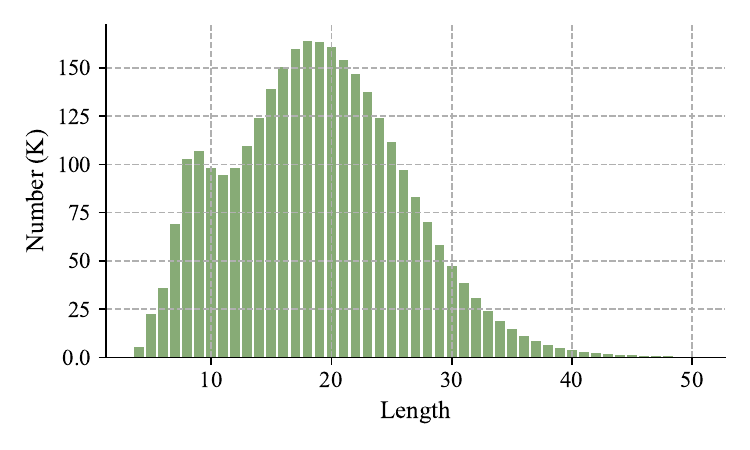}
  \end{minipage}
  }

  \vspace{0.25em} 
    \centering
    \subfigure[Multi-round instruction.]{
  \begin{minipage}[b]{0.48\linewidth}
    \centering
    \includegraphics[width=\linewidth]{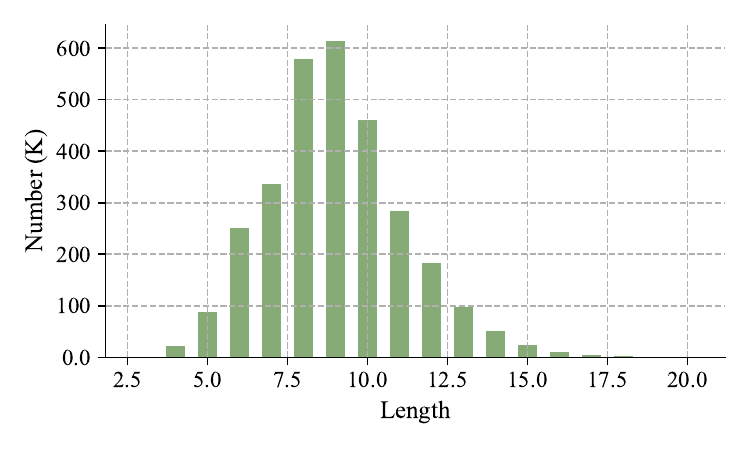}
  \end{minipage}
  }
   \hspace{0.05cm}
   \subfigure[Multi-round response.]{
  \begin{minipage}[b]{0.48\linewidth}
    \centering
    \includegraphics[width=\linewidth]{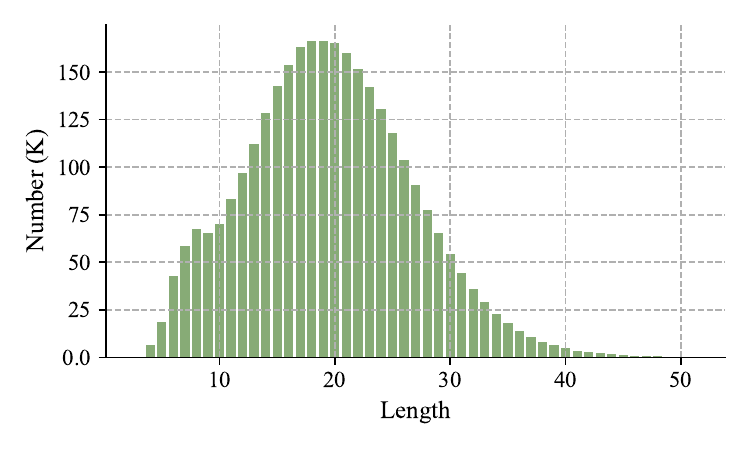}
  \end{minipage}
  }

  \caption{\textbf{Sentence length of all types of data in our T3D dataset.}}
  \label{fig:abla_sentence_len}

\end{figure*}

%% file: Appendix/image/Qwen_generation_data.tex
 \begin{figure*}[t]
\centering
  \includegraphics[width=\linewidth]{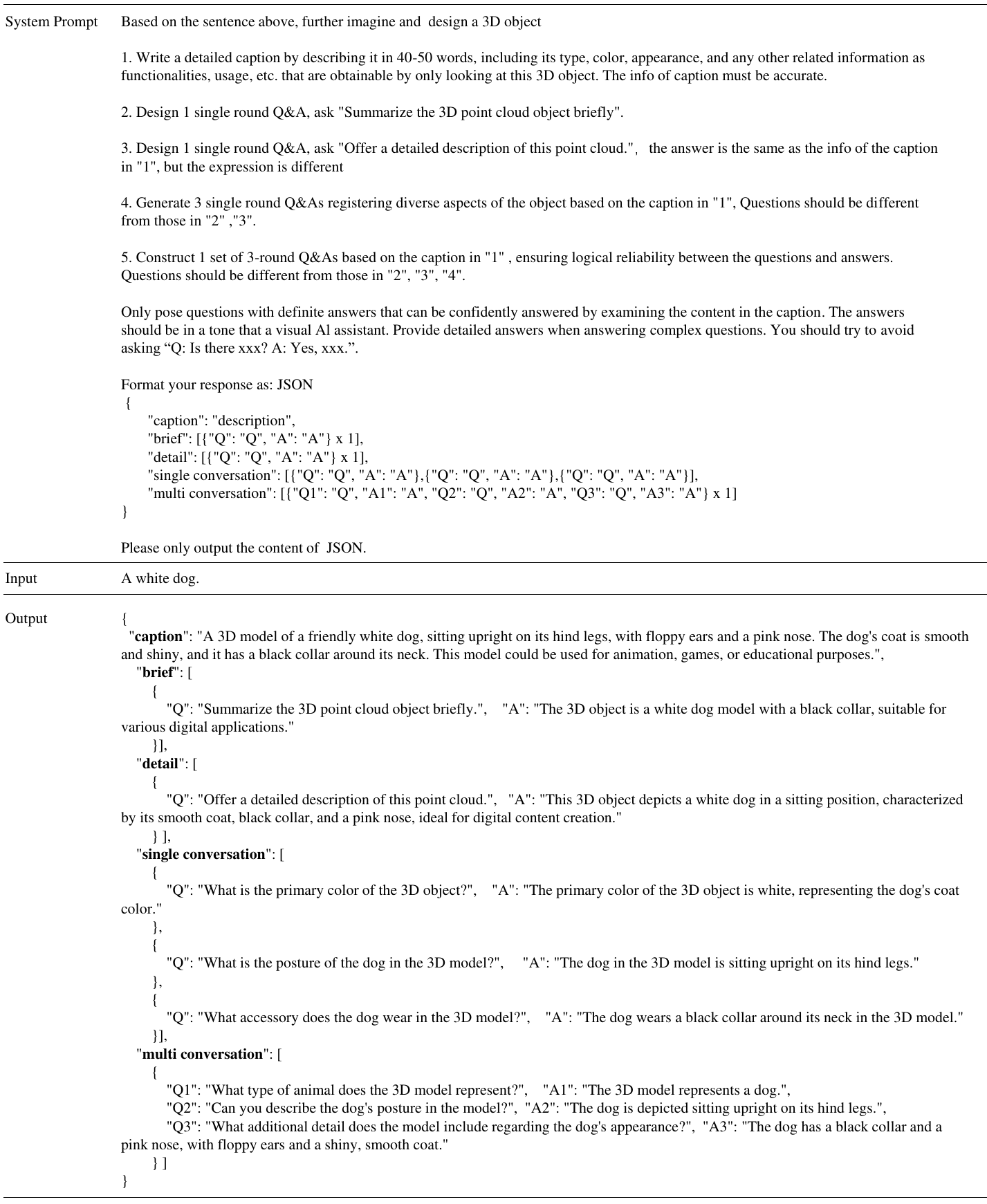}
  \caption{\textbf{An example of data generation pipeline using Qwen2-72B-Instruct.}
Given any object category, the LLM generates 5 types of data based on our designed prompt templates. 
The output is in JSON format, including a caption, brief description, detailed description, three rounds of single-turn conversation, and one round of multi-turn conversation.
}
 
  \label{fig:abla_Qwen_generation}
\end{figure*}

%% file: Appendix/table/instruct_list.tex
\begin{table*}[t]
\caption{\textbf{The instruction list for brief descriptions.} 
We follow PointLLM~\cite{xu2023pointllm} by using more diverse instructions, replacing the generated simpler questions in brief descriptions, as the final instructions.
}
\label{tab:abla_brief_des_prompt}
\begin{tabular}{@{}p{\linewidth}@{}} 
\toprule
\begin{itemize}
\item Summarize the 3D point cloud object briefly.
\item What kind of object is depicted by this point cloud?
\item Provide a short explanation of this 3D structure.
\item What does this collection of points represent?
\item Offer a succinct summary of this 3D object.
\item Can you give a brief overview of this point cloud?
\item Characterize the object this point cloud is illustrating.
\item Share a brief interpretation of this 3D point cloud.
\item Provide an outline of this 3D shape's characteristics.
\item What object is this point cloud rendering?
\item Deliver a quick description of the object represented here.
\item How would you describe the 3D form shown in this point cloud?
\item What is the nature of the object this point cloud is representing?
\item Present a compact account of this 3D object's key features.
\item What can you infer about the object from this point cloud?
\item Offer a clear and concise description of this point cloud object.
\item How would you summarize this 3D data set?
\item Give a brief explanation of the object that this cloud of points forms.
\item What kind of structure does this 3D point cloud depict?
\item Could you delineate the form indicated by this point cloud?
\item Express in brief, what this point cloud is representing.
\item Give a quick overview of the object represented by this 3D cloud.
\item Convey a summary of the 3D structure represented in this point cloud.
\item What kind of object is illustrated by this collection of points?
\item Describe the object that this point cloud forms.
\item How would you interpret this 3D point cloud?
\item Can you briefly outline the shape represented by these points?
\item Give a concise interpretation of the 3D data presented here.
\item Explain the object this point cloud depicts succinctly.
\item Offer a summary of the 3D object illustrated by this cloud.
\end{itemize} \\ \bottomrule
\end{tabular}
\label{fig:abla_brief_inst}
\end{table*}

\begin{table*}
\caption{\textbf{The instruction list for detailed descriptions.} 
We follow PointLLM~\cite{xu2023pointllm} by using more diverse instructions, replacing the generated simpler questions in detailed descriptions, as the final instructions.
}
\label{tab:abla_detailed_des_prompt}
\begin{tabular}{@{}p{\linewidth}@{}} 
\toprule
\begin{itemize}
\item Can you tell me more about this?
\item What does this represent?
\item Can you describe this in more detail?
\item I'm interested in this, can you explain?
\item What is this object made of?
\item Could you provide more info about this?
\item What exactly am I looking at here?
\item What is this?
\item Could you describe the detailed structure of this?
\item This looks interesting, can you expand on it?
\item Can you explain more about this form?
\item What can you tell me about the shape of this object?
\item Could you delve deeper into this?
\item I want to know more about this, can you help?
\item Can you walk me through the details of this object?
\item Can you provide a comprehensive account of this object?
\item Offer a detailed interpretation of this point cloud.
\item Please elucidate on the characteristics of this form.
\item Could you provide an in-depth description of this structure?
\item What does this cloud represent in its entirety?
\item Elaborate on the details of this point cloud, please.
\item Kindly furnish me with more information about this object.
\item Please expand on the intricate structure of this form.
\item Provide a meticulous explanation of what these points represent.
\item I request a detailed breakdown of this structure.
\item Give a thorough rundown of this point cloud.
\item Can you offer a complete analysis of this object?
\item I would like a comprehensive explanation of this form.
\item Please detail the specific features of this point cloud.
\item Could you elaborate extensively on what this represents?
\end{itemize} \\ \bottomrule
\end{tabular}
\label{fig:abla_detail_inst}
\end{table*}

%% file: Appendix/image/sample_brief.tex
\begin{figure*}[t]
    \centering
  \includegraphics[width=\textwidth]{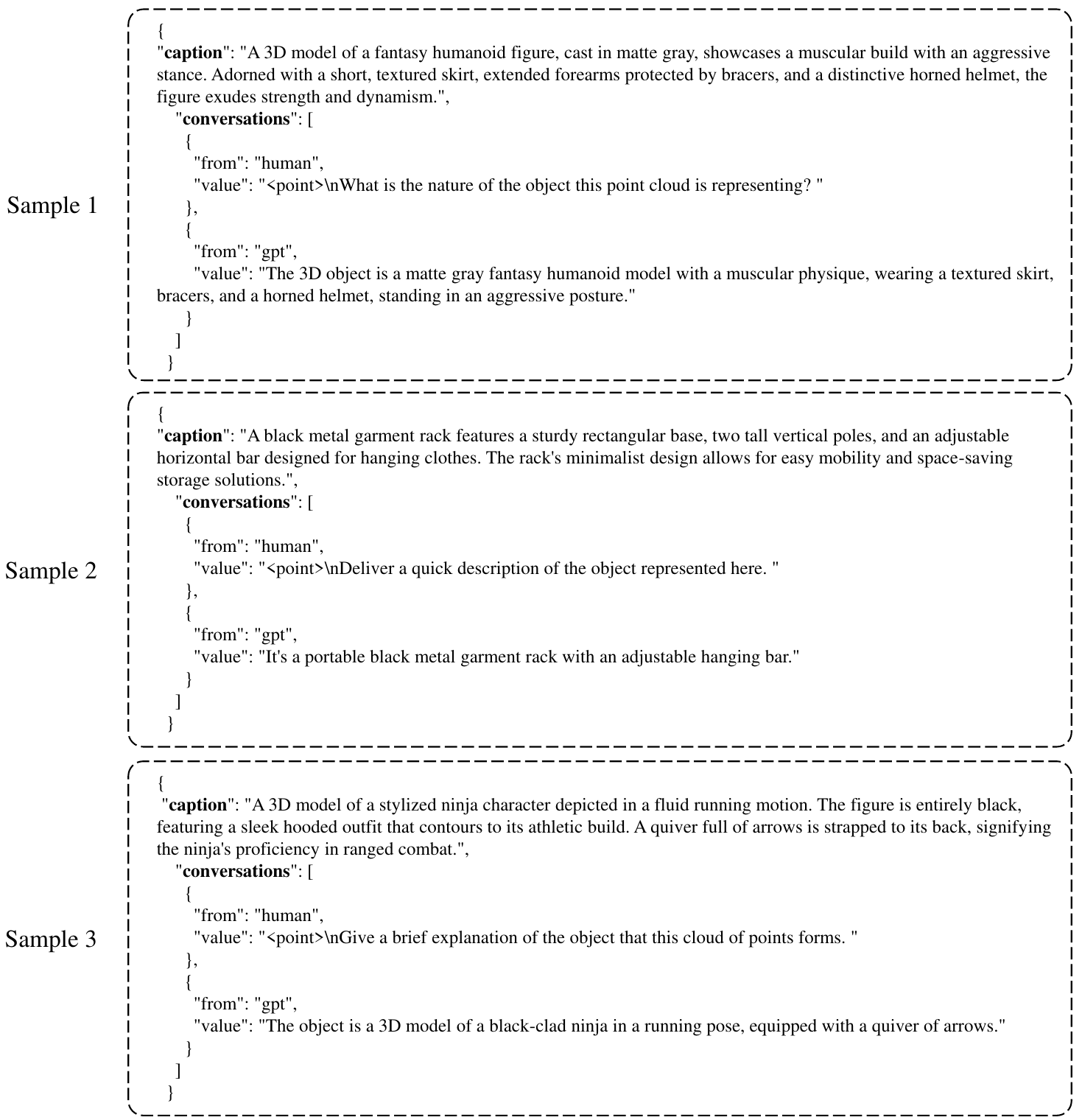}
  \caption{
    \textbf{Brief description samples of our T3D dataset.}
    }
  \label{fig:abla_sample_brief}
  \vspace{-5pt}
\end{figure*}

%% file: Appendix/image/sample_detail.tex
\begin{figure*}[t]
    \centering
  \includegraphics[width=\textwidth]{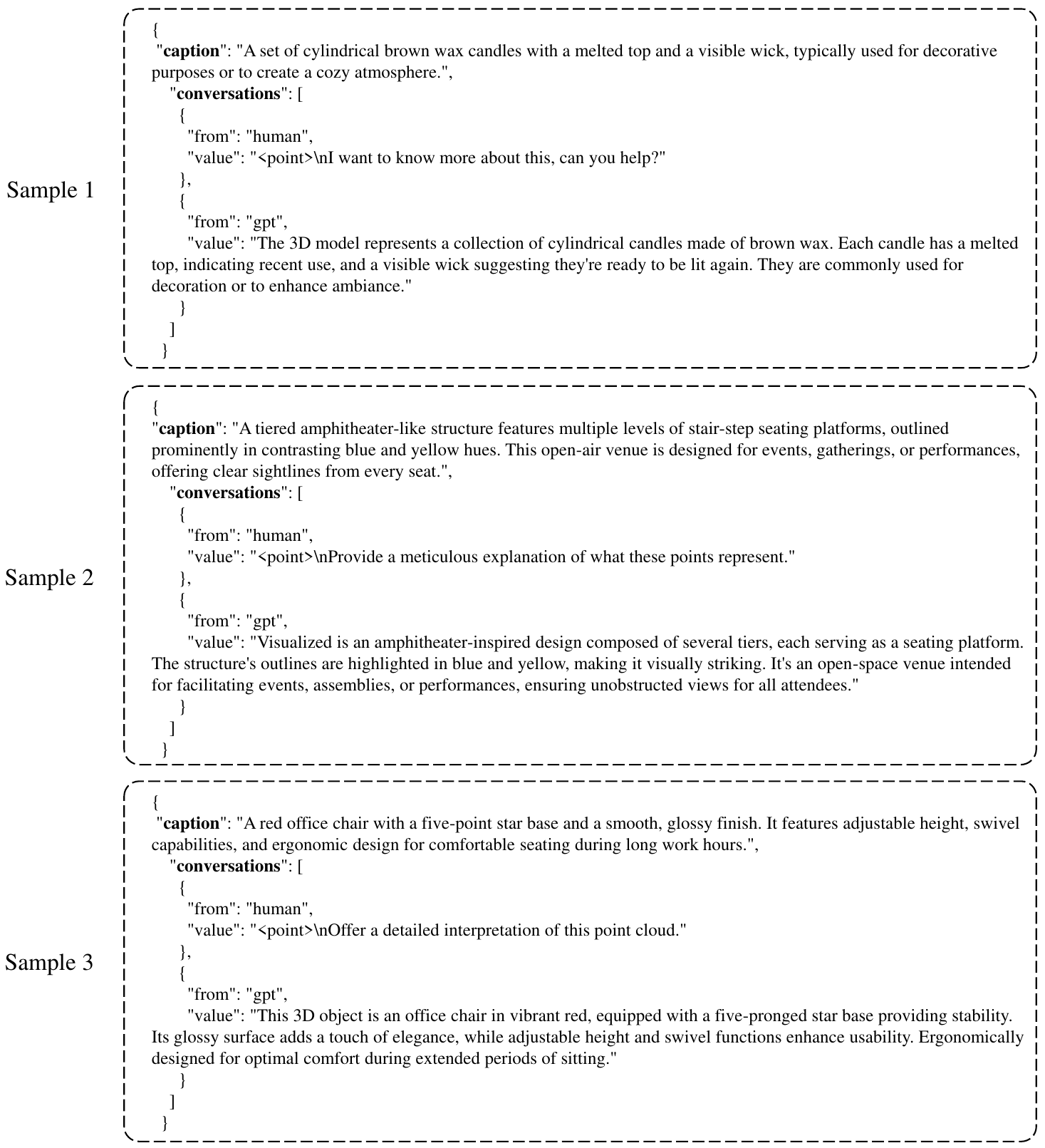}
  \caption{
    \textbf{Detailed description samples of our T3D dataset.}
    }
  \label{fig:abla_sample_detail}
  \vspace{-5pt}
\end{figure*}

%% file: Appendix/image/sample_single_conv.tex
\begin{figure*}[t]
    \centering
  \includegraphics[width=\textwidth]{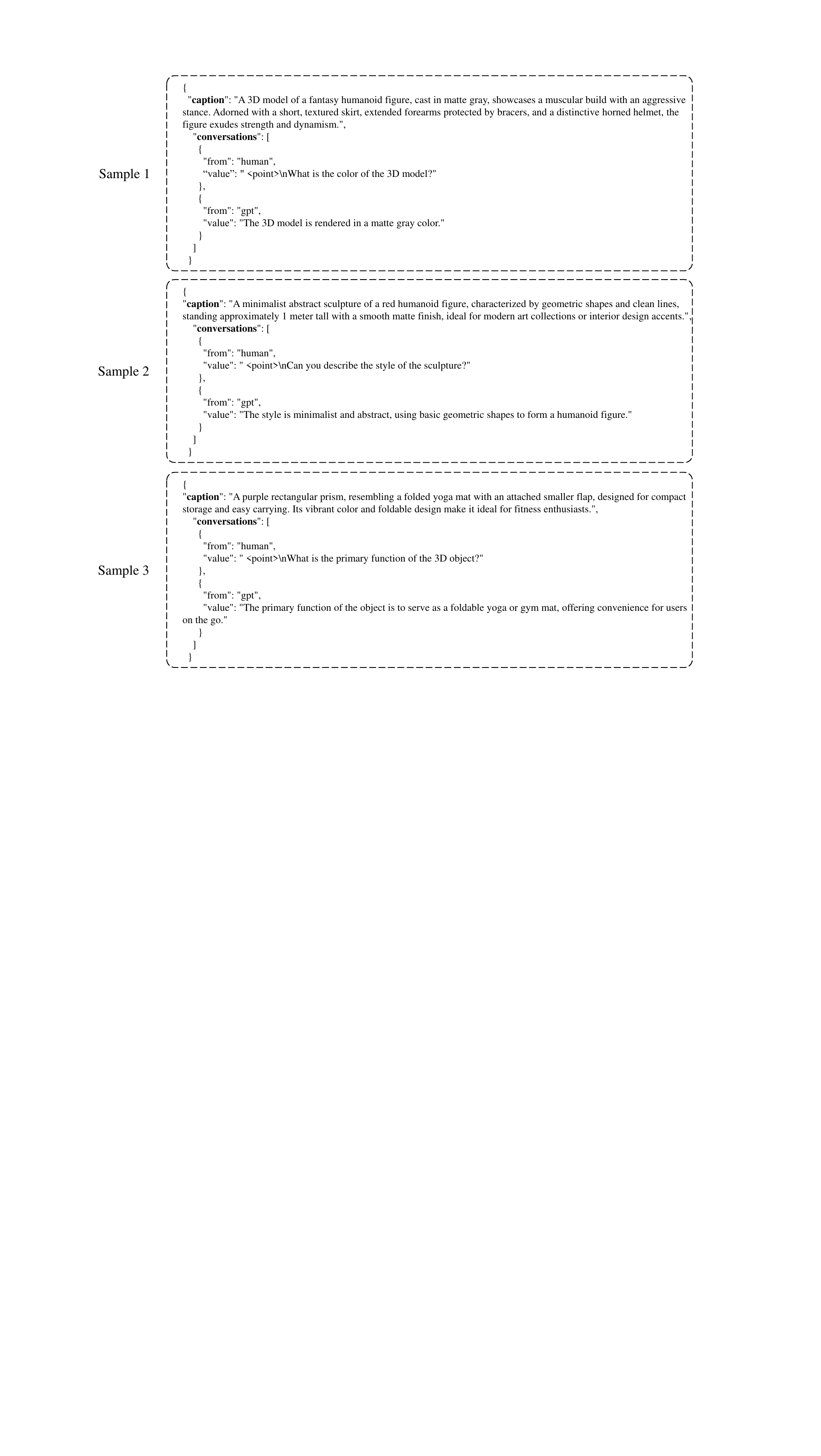}
  \caption{
    \textbf{Single-round conversation samples of our T3D dataset.}
    }
  \label{fig:abla_sample_single_conv}
  \vspace{-5pt}
\end{figure*}

%% file: Appendix/image/sample_multi_conv.tex
\begin{figure*}[t]
    \centering
  \includegraphics[width=\textwidth]{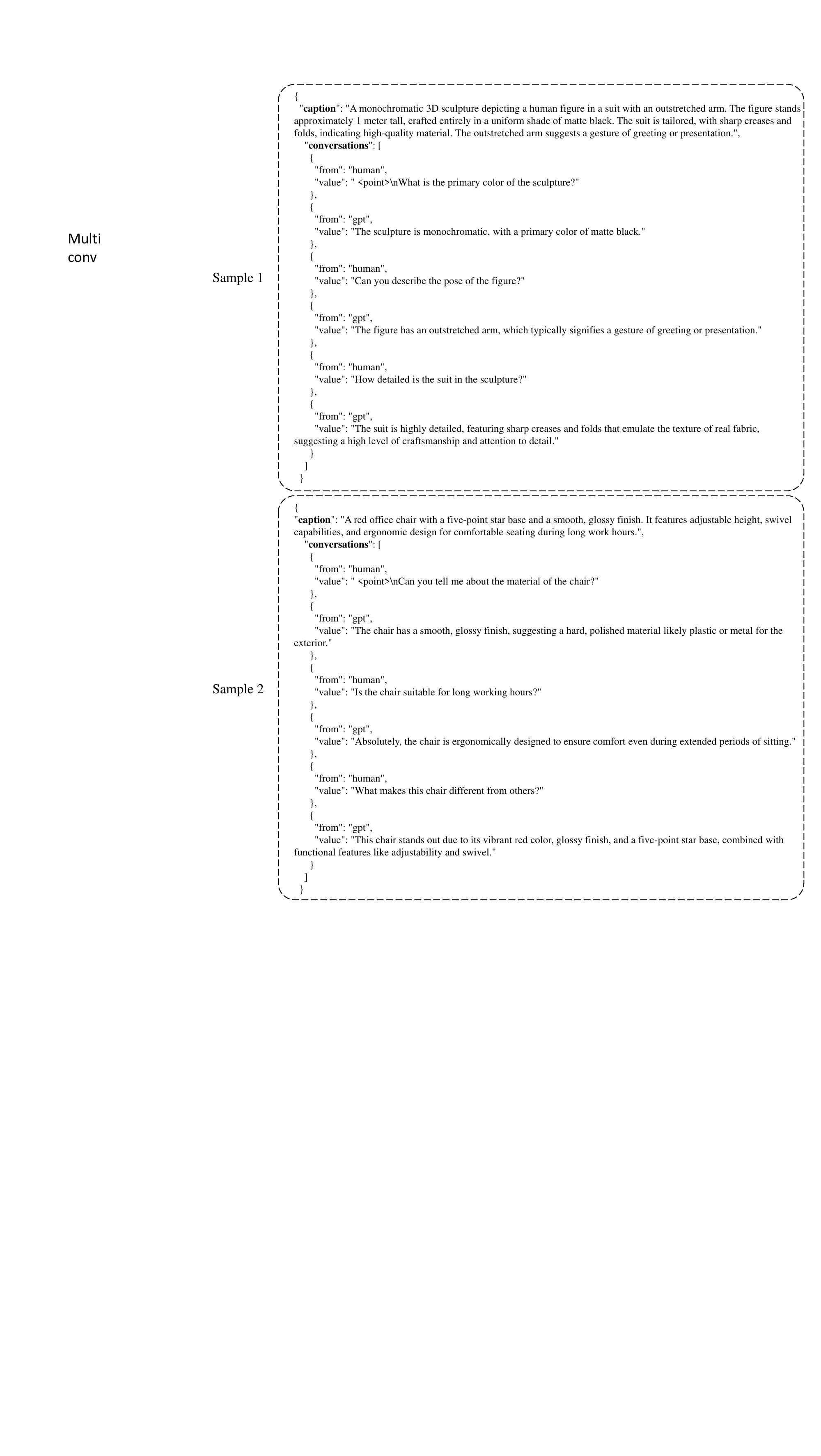}
  \caption{
    \textbf{Multi-round conversation samples of our T3D dataset.}
    }
  \label{fig:abla_sample_multi_conv}
  \vspace{-5pt}
\end{figure*}

%% file: Appendix/image/conv_text_only.tex
 \begin{figure*}[t]
\centering
  \includegraphics[width=0.8\linewidth]{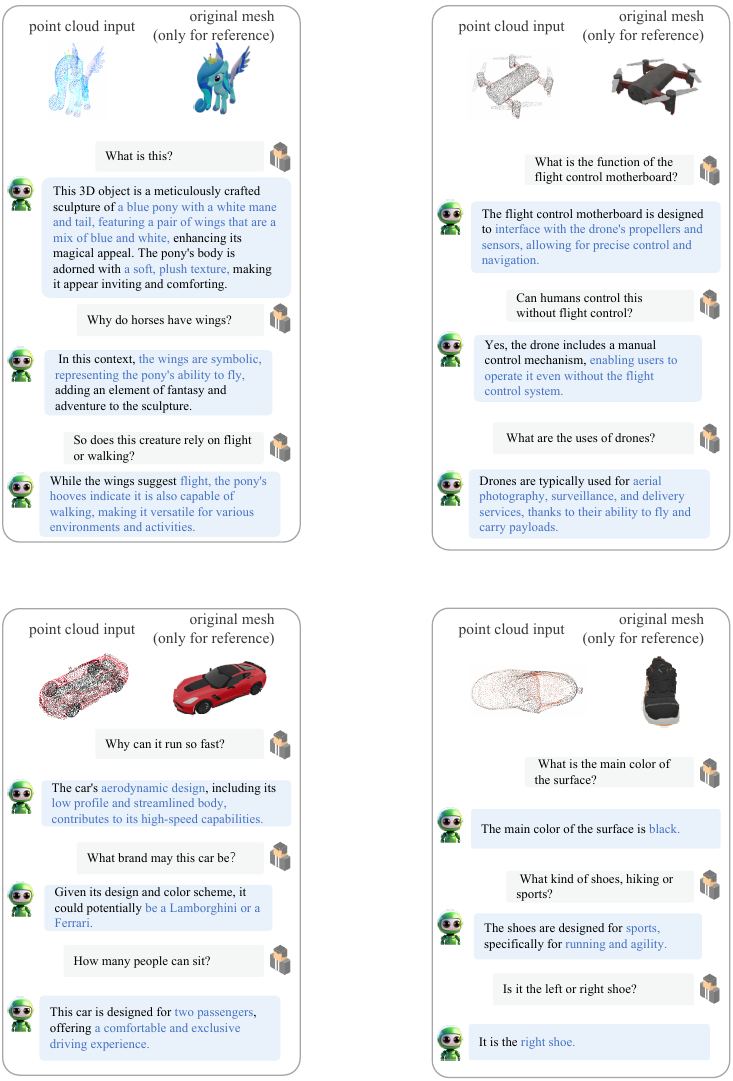}
  \caption{\textbf{Conversation examples of GreenPLM-0 (text-only data).}
}
 
  \label{fig:abla_conv_text_only}
\end{figure*}

%% file: Appendix/image/conv_stage_3.tex
 \begin{figure*}[t]
\centering
  \includegraphics[width=0.8\linewidth]{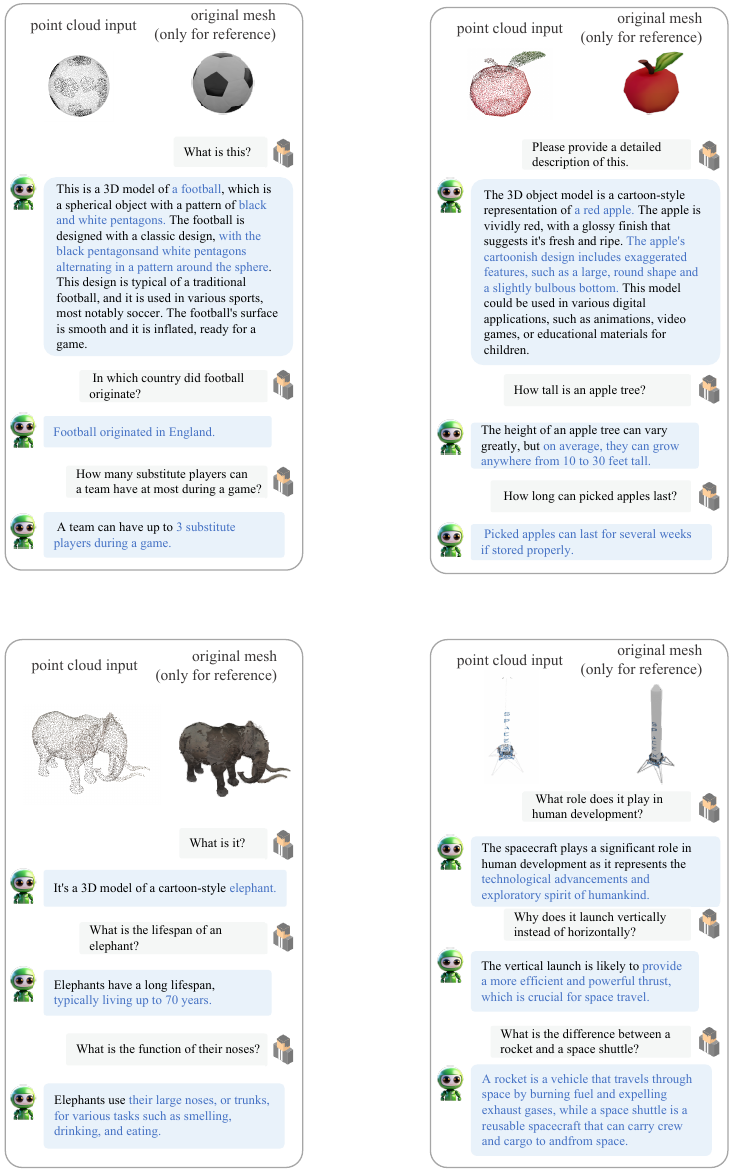}
  \caption{\textbf{Conversation examples of GreenPLM (limited 3D data).}
}
 
  \label{fig:abla_conv_stage_3}
\end{figure*}

%% file: Appendix/table/setting.tex
\begin{table*}[t]
  \centering

  \caption{\textbf{Detailed training settings.}
  %
   }

  \resizebox*{\linewidth}{!}{
\begin{tabular}{cc|ccc}
\toprule
\multicolumn{2}{c|}{Setting}                                          & Stage I                                                                                   & Stage II                                                                                  & Stage III                                                                                           \\ \midrule
\multirow{3}{*}{Dataset}                        & Dataset Source     & \multicolumn{2}{c}{T3D (Ours)}                                                                                                                                                        & Point-text Instruction Dataset~\cite{xu2023pointllm}                                                                      \\ \addlinespace[2pt] \cline{2-5}  \addlinespace[2pt]
                                                & Dataset Type       & Brief Description                                                                         & \begin{tabular}[c]{@{}c@{}}Detail Description\\ \& Conversation\end{tabular}              & \begin{tabular}[c]{@{}c@{}}Brief Description\\ \& Detail Description\\ \& Conversation\end{tabular} \\ \addlinespace[2pt]\cline{2-5} \addlinespace[2pt]
                                                & Dataset Scale      & 1M                                                                                        & 210K                                                                                      & 90K                                                                                                 \\ \midrule
\multicolumn{2}{c|}{Training Time}                                    & 12.6 Hours                                                                                & 5.9 Hours                                                                                 & 8.1 Hours                                                                                           \\ \midrule
\multicolumn{2}{c|}{Trainable Parameters}                             & 12.6M                                                                                     & 62.9M                                                                                     & 63.3M                                                                                               \\ \midrule
\multicolumn{2}{c|}{Batch Size}                                       & 16                                                                                        & 14                                                                                        & 25                                                                                                  \\ \midrule
\multicolumn{2}{c|}{Epoch}                                            & 1                                                                                         & 1                                                                                         & 3                                                                                                   \\ \midrule
\multicolumn{2}{c|}{Learning Rate}                                    & 1e-3                                                                                      & 2e-4                                                                                     & 5e-5                                                                                               \\ \midrule
\multicolumn{2}{c|}{Noise Std}                                        & 0.05                                                                                      & 0.05                                                                                      & -                                                                                                   \\ \midrule
\multirow{4}{*}{Text Encoder}                   & Parameters         & 5B                                                                                        & 5B                                                                                        & -                                                                                                   \\
                                                & Hidden Size        & 1280                                                                                      & 1280                                                                                      & -                                                                                                   \\
                                                & Head of Attention  & 20                                                                                        & 20                                                                                        & -                                                                                                   \\
                                                & Number of Layer    & 32                                                                                        & 32                                                                                        & -                                                                                                   \\ \midrule
\multirow{7}{*}{Point Cloud Encoder}            & Parameters         & -                                                                                         & -                                                                                         & 22.6M                                                                                               \\
                                                & Point Number       & -                                                                                         & -                                                                                         & 8192                                                                                                \\
                                                & Point Group Size   & -                                                                                         & -                                                                                         & 64                                                                                                  \\
                                                & Point Patch Number & -                                                                                         & -                                                                                         & 512                                                                                                 \\
                                                & Hidden Size        & -                                                                                         & -                                                                                         & 384                                                                                                 \\
                                                & Head of Attention  & -                                                                                         & -                                                                                         & 6                                                                                                   \\
                                                & Number of Layer    & -                                                                                         & -                                                                                         & 12                                                                                                  \\ \midrule
\multirow{2}{*}{0M-Pooling}                     & FPS Token Number   & -                                                                                         & -                                                                                         & 32                                                                                                  \\
                                                & KNN Token Number   & -                                                                                         & -                                                                                         & 8                                                                                                   \\ \midrule
\multirow{2}{*}{Projector MLP}                  & Number of Layer    & 2                                                                                         & 2                                                                                         & 2                                                                                                   \\
                                                & Dimension          & \begin{tabular}[c]{@{}c@{}}1024~-\textgreater~{}3072\\ 3072~-\textgreater~{}3072\end{tabular} & \begin{tabular}[c]{@{}c@{}}1024~-\textgreater~{}3072\\ 3072~-\textgreater~{}3072\end{tabular} & \begin{tabular}[c]{@{}c@{}}1024~-\textgreater~{}3072\\ 3072~-\textgreater~{}3072\end{tabular}           \\ \midrule
\multirow{6}{*}{Large Lanuguage Model Backbone} & Parameters         & 3.8B                                                                                      & 3.8B                                                                                      & 3.8B                                                                                                \\
                                                & Rank of LoRA       & -                                                                                         & 32                                                                                        & 32                                                                                                  \\
                                                & Alpha of LoRA      & -                                                                                         & 64                                                                                        & 64                                                                                                  \\
                                                & Number of Layer    & 32                                                                                        & 32                                                                                        & 32                                                                                                  \\
                                                & Head of Attention  & 32                                                                                        & 32                                                                                        & 32                                                                                                  \\
                                                & Hidden Size        & 3072                                                                                      & 3072                                                                                      & 3072                                                                                                \\ \bottomrule
\end{tabular}}
\label{tb:abla_train_setting}
\end{table*}

%% file: Appendix/image/tain_inference.tex
\begin{figure*}[t]
    \centering
  \includegraphics[width=\textwidth]{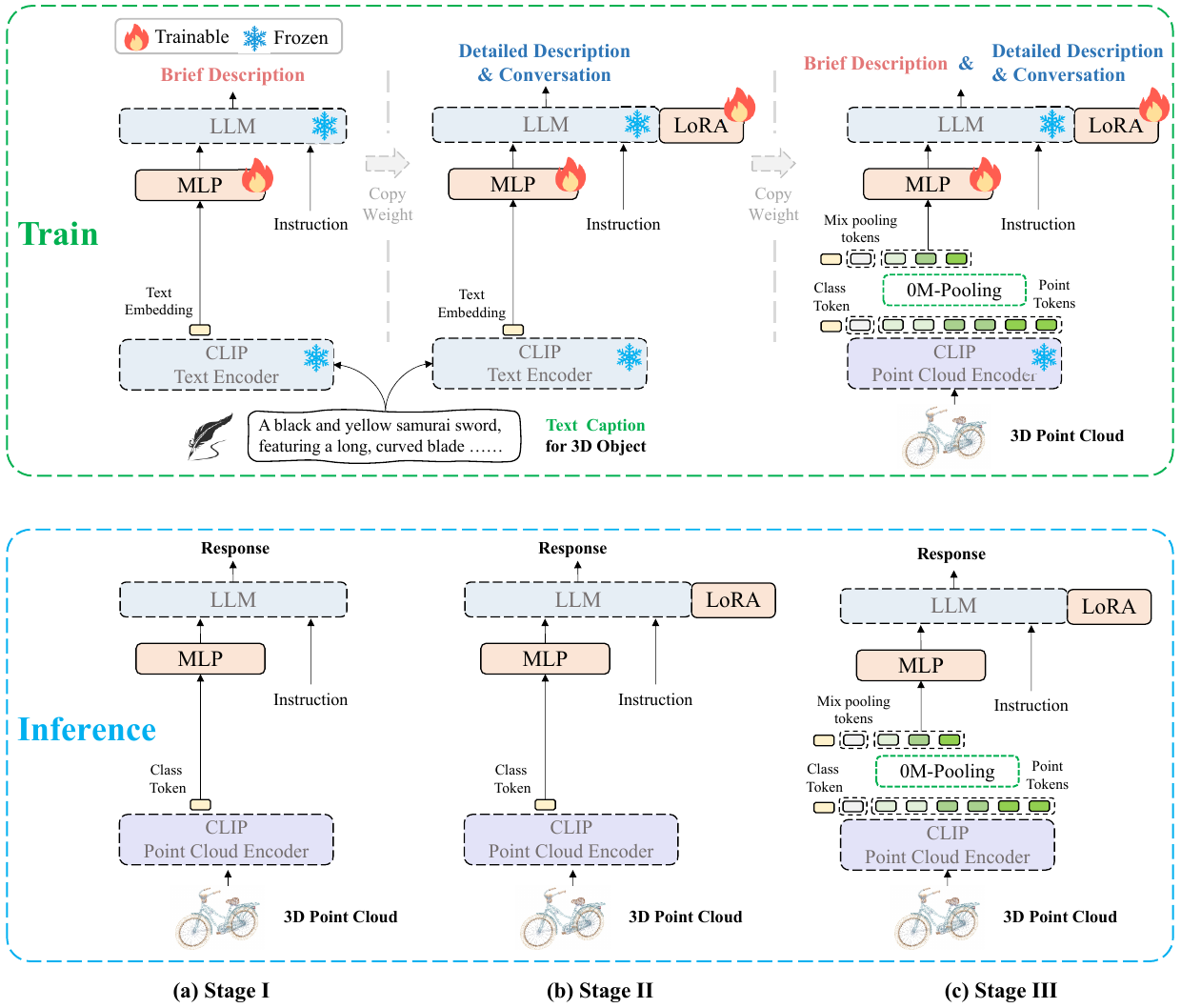}
  \caption{
    \textbf{Architectures of training and inferencing in three stages.}
    }
  \label{fig:abla_train_inference}
  \vspace{-5pt}
\end{figure*}